\documentclass[runningheads]{llncs}

\usepackage{eccv}

\usepackage{eccvabbrv}

\usepackage{graphicx}
\usepackage{booktabs}
\usepackage{multirow}
\usepackage{pifont}
\usepackage{colortbl}
\usepackage{enumitem}
\usepackage{bm}
\usepackage{tabularx}
\usepackage{import}

\usepackage[accsupp]{axessibility}  

\usepackage{hyperref}

\usepackage{orcidlink}

\definecolor{cvprblue}{rgb}{0.21,0.49,0.74}
\definecolor{mediumgray}{gray}{0.5}
\newcommand{\xmark}{\ding{55}}
\newcommand{\cmark}{\ding{51}} 

\newcommand{\equalcontrib}{\textsuperscript{*}}

\begin{document}

\title{ContextFlow: In-Context Flow Matching for Robot Manipulation} 

\titlerunning{ContextFlow}

\author{Jian Ding\inst{1}\orcidlink{0000-0002-7188-5884}\equalcontrib \and Xianjie Dai\inst{1}\orcidlink{0009-0006-2842-611X}\equalcontrib \and
Roei Herzig\inst{2}\orcidlink{0000-0002-5451-6059} \and Nussair Hroub\inst{1}\orcidlink{0009-0004-4346-4208} \and Jinjie Mai\inst{1}\orcidlink{0000-0002-3396-1970} \and Dengxin Dai\inst{3}\orcidlink{0000-0001-5440-9678} \and Bernard Ghanem\inst{1}\orcidlink{0000-0002-5534-587X} \and Mohamed Elhoseiny\inst{1}\orcidlink{0000-0001-9659-1551}\textsuperscript{\dag}  }

\authorrunning{J.~Ding, X.~Dai et al.}

\institute{KAUST, Thuwal, Saudi Arabia; \email{\{jian.ding, dai.xianjie, nussair.hroub, jinjie.mai, bernard.ghanem, mohamed.elhoseiny\}@kaust.edu.sa}  \and University of California, Berkeley, USA; \email{roeiherz@gmail.com} \and Zurich, Switzerland; \email{ddx2004@gmail.com}}

\maketitle

\begingroup
\renewcommand{\thefootnote}{*}
\footnotetext{Equal contribution.}
\endgroup

\begingroup 
\renewcommand{\thefootnote}{\dag} 
\footnotetext{Corresponding author.} 
\endgroup

\begin{abstract}
Although highly effective in vision and language domains, applying in-context learning to robotics remains challenging. Existing autoregressive in-context imitation methods discretize continuous actions and exacerbate the accumulation of early prediction errors through next-token prediction, limiting their generalization on unseen task configurations. Meanwhile, flow-matching policies have been explored for continuous robot control and can help mitigate compounding errors; however, \textit{in-context imitation learning within a flow-matching framework remains underexplored}. 
To address these limitations, we introduce ContextFlow, a \textit{conditional flow-matching model} that learns continuous action distributions for in-context imitation learning. ContextFlow conditions flow-based action prediction on demonstrations and observations, enabling robust generation from noisy action distributions. To better encode multimodal in-context demonstrations, we adapt perceiver-style \textit{multimodal context compressors} that distill visual, proprioceptive, and action sequences into compact, task-relevant latent representations.
On LIBERO, ContextFlow outperforms ICRT by \textbf{35 percentage points} in average success rate on unseen task configurations, while matching the performance of the task-specific fine-tuned VLA model $\pi_0$ without any fine-tuning on unseen tasks. On real robots, it generalizes to unseen configurations of both single-arm and bimanual tasks, achieving \textbf{40\%} success on a new pen-uncapping configuration. Project Page: \url{https://dingjiansw101.github.io/contextflow-page/}.


\end{abstract}
\section{Introduction}
\label{sec:intro}

\begin{figure}[t]
\centering
\includegraphics[width=0.98\linewidth]{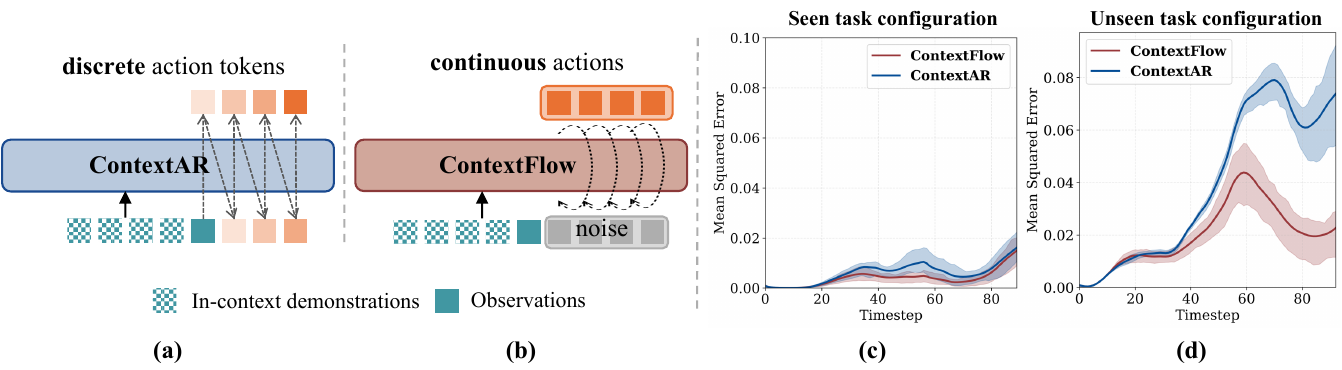}
\caption{\textbf{Autoregressive (a) vs.\ Flow Matching (b) for In-Context Imitation Learning on action decoding and compounding errors.}
\textit{In-context imitation learning} refers to performing a task conditioned on in-context demonstrations without updating model parameters. (a) In-context autoregressive models generate discretized actions step by step, which leads to compounding errors, especially under task distribution shift. (b) Our proposed in-context flow matching framework instead predicts a chunk of continuous actions, alleviating error accumulation and improving generalization. (c) Comparison of compounding errors under \textit{seen} task configuration between autoregressive and flow-matching approaches. (d) Comparison of compounding errors under \textit{unseen} task configuration.}
    \label{fig:overview}
\end{figure}

A hallmark of human intelligence is the ability to solve a new task after observing only a few demonstrations. In machine learning, this capability is often referred to as \textit{in-context learning}: a model is conditioned at test time on a small set of input-output examples and adapts to a new task without gradient updates. In-context learning has been prominently demonstrated in NLP: Large language models (LLMs) such as GPT-3~\cite{gpt3}, trained with autoregressive objectives on massive text corpora, exhibit in-context learning as an emergent behavior~\cite{wang2023learning,min2022rethinking}.

Similar paradigms have been explored in robotics as \textit{in-context imitation learning}, where demonstrations are represented as sequences of observations and expert actions. This promises to adapt a policy to a new task from only a few demonstrations, avoiding fine-tuning that existing imitation learning methods~\cite{osa2018algorithmic,alvinn,ross2011reduction} require. Yet generalization in robotic manipulation is hard, as policies must produce continuous actions over long horizons from few demonstrations. Consequently, prior in-context imitation work~\cite{icrt,shah2025mimicdroid}, like ours, targets \textit{unseen configurations within related primitives} (\textit{e.g.}, unseen objects or spatial arrangements) rather than arbitrary new tasks. Many existing in-context imitation learning approaches~\cite{icrt,actiontoken,shah2025mimicdroid} instantiate the policy as an \textit{autoregressive} model (ARM)~\cite{touvron2023llama,gpt3}, a well-studied and scalable structure that can be directly adapted from LLMs. However, ARM-based policies (Fig.~\ref{fig:overview} (a)) face two fundamental limitations in robotic control. First, they predict actions sequentially, with each future token conditioned on previously predicted ones; as a result, \textit{an early prediction error can propagate and compound over time}. Although this issue can be partially mitigated when training and evaluation are drawn from the \textit{same} task configurations (see Fig.~\ref{fig:overview} (c)), it becomes much more pronounced in in-context imitation learning, where training and evaluation are drawn from \textit{different} task configurations (see Fig.~\ref{fig:overview} (d)). Under such a distribution shift, early errors occur more frequently and are amplified throughout the autoregressive rollout. Second, many ARM-based policies rely on action tokenizers~\cite{actiontoken,pertsch2025fast} that discretize continuous actions, introducing \textit{reconstruction error}~\cite{pertsch2025fast,actiontoken} and further exacerbating execution failures.

In the broader imitation learning literature, action chunking~\cite{kim2025fine,zhao2023learning} or continuous action modeling, such as flow matching~\cite{flowmatchtutorial,lipman2022flow,liu2022rectified,pi0}, has been explored to alleviate reconstruction and compounding errors. Flow matching (see Fig.~\ref{fig:overview} (b)) operates directly in continuous action space and predicts an \textit{action chunk in parallel} rather than \textit{token by token}, which helps mitigate compounding errors, especially under distribution shift. However, unlike autoregressive in-context learning, whose architecture and training recipe are well-established in LLMs, \textit{the incorporation of in-context learning into a flow-matching framework remains underexplored}. In this paper, we build on the architectural principles of a vision--language--action flow model, $\pi_0$~\cite{pi0}, and extend it to support multimodal in-context imitation learning.
Concretely, we begin by building \textit{ContextFlow-Plain}, which uses a \textit{context expert} to encode the in-context demonstrations, along with the current observation, into a set of context tokens. An \textit{action expert} then takes the current proprioceptive state and noisy action tokens, and refines them through a sequence of flow-matching updates while attending to the context tokens, ultimately predicting a clean \textit{continuous} action chunk. By generating continuous action vectors \textit{in parallel} and avoiding action discretization, ContextFlow-Plain mitigates the compounding errors that are especially harmful under unseen task configurations.

Although ContextFlow-Plain has already equipped a flow-matching framework with in-context learning ability, its performance is further constrained by the cost of encoding long, multimodal in-context demonstrations with a large vision--language--action encoder. Robot trajectories often span hundreds of timesteps and include camera observations, proprioception, and actions, all of which exhibit substantial spatial and temporal redundancy; processing these raw sequences directly with attention is prohibitively expensive. To address this efficiency bottleneck, we propose \textit{ContextFlow}, which augments ContextFlow-Plain with \textit{multimodal context compressors} adapted from perceiver-style architectures~\cite{perceiver,perceiverio,qformer}. We tailor this structure to in-context demonstration encoding by distilling visual, proprioceptive, and action streams into a compact, fixed-size set of latent context tokens, preserving task-relevant information while substantially reducing the cost of context encoding and cross-attention.

We demonstrate the effectiveness of ContextFlow on the LIBERO simulation. To further evaluate ContextFlow on diverse and challenging real-world robot tasks, we collect a new dataset of expert demonstrations on the ALOHA platform. The dataset contains 1,318 multimodal trajectories across 25 task configurations, spanning both single-arm skills and challenging bimanual tasks that require high-frequency control and closed-loop feedback.
Our contributions are three-fold:
\begin{itemize}
\item We formulate in-context imitation learning for robotics within a \textit{conditional flow matching} framework, and empirically observe lower errors on unseen configurations.
\item We curate a real-world ALOHA dataset for in-context imitation learning, featuring both single-arm tasks and fine-grained \textit{bimanual} manipulation tasks. 
\item We evaluate ContextFlow both in LIBERO simulation and on ALOHA real robots. Across these settings, ContextFlow outperforms autoregressive in-context imitation learning baselines in our evaluations, while matching the performance of task-specific, fine-tuned large vision--language--action models without requiring any fine-tuning on unseen configurations.

\end{itemize}

\section{Related Work}
\label{related_work}

\noindent \textbf{In-Context Imitation Learning.}
\textit{In-context learning} enables adaptation from a few examples \textit{without weight updates}, a capability popularized by autoregressive LLMs trained with next-token prediction~\cite{gpt3,gpt4,llama,qwen}. In robotics, KAT~\cite{dipalo2024kat} and RoboPrompt~\cite{roboprompt} perform few-shot imitation by tokenizing observations/actions and querying external LLMs, while earlier work~\cite{duan2017one} and Prompt-DT~\cite{xu2022prompting} rely on full-state observations. Instant Policy~\cite{vosylius2024instant} and IMOP~\cite{zhang2024one} use 3D structures, requiring external segmentation or IK execution; the latter can fail near singularities in fine manipulation. In contrast, ContextFlow uses only RGB and directly outputs executable actions. Similar to ours, ICRT~\cite{icrt} and LipVQ-VAE~\cite{actiontoken} train end-to-end on RGB observations using autoregressive action tokens, with LipVQ-VAE improving tokenization, and RICL~\cite{sridhar2025ricl} extends autoregressive VLA models with in-context adaptability. We instead formulate in-context imitation learning via conditional flow matching, modeling continuous action chunks directly without action discretization. We further demonstrate real-world \emph{fine-grained bimanual} manipulation requiring \textit{high-frequency} control and \textit{closed-loop} feedback, a setting still underexplored in prior work.

\noindent \textbf{Vision Language Action Models.} Vision language action (VLA) models extend vision language models (VLMs) to predict low-level robotic actions from visual observations and language instructions. RT-1~\cite{brohan2022rt}, RT-2~\cite{brohan2023rt}, and OpenVLA~\cite{kim2024openvla} adopt autoregressive architectures trained with a next-token prediction objective over discrete action tokens. To improve token efficiency and reduce discretization artifacts, Fast Tokenizer~\cite{pertsch2025fast} proposes more compact action vocabularies, and OpenVLA-OFT~\cite{kim2025fine} shows that adding a parallel continuous-action prediction head can further boost performance. More recent VLA models~\cite{wen2025tinyvla,liu2024rdt,pi0,pi05} incorporate diffusion~\cite{li2022diffusion} or flow-matching~\cite{flowmatchtutorial} heads for continuous control, with $\pi_0$~\cite{pi0,pi05} demonstrating strong instruction-following capabilities. However, these models primarily rely on textual instructions, which limit their generalization ability to new task configurations. In contrast, our approach uses \emph{multimodal in-context demonstrations} as the task prompt. VIMA~\cite{jiang2023vima} also studies multimodal in-context prompting for robotics, but predicts only \textit{2--5 high-level} actions (e.g., ``pick and place'' or ``wipe'') per episode, whereas our model predicts hundreds of low-level action sequences.

\begin{figure*}[!t]
\centering
\includegraphics[width=0.98\linewidth]{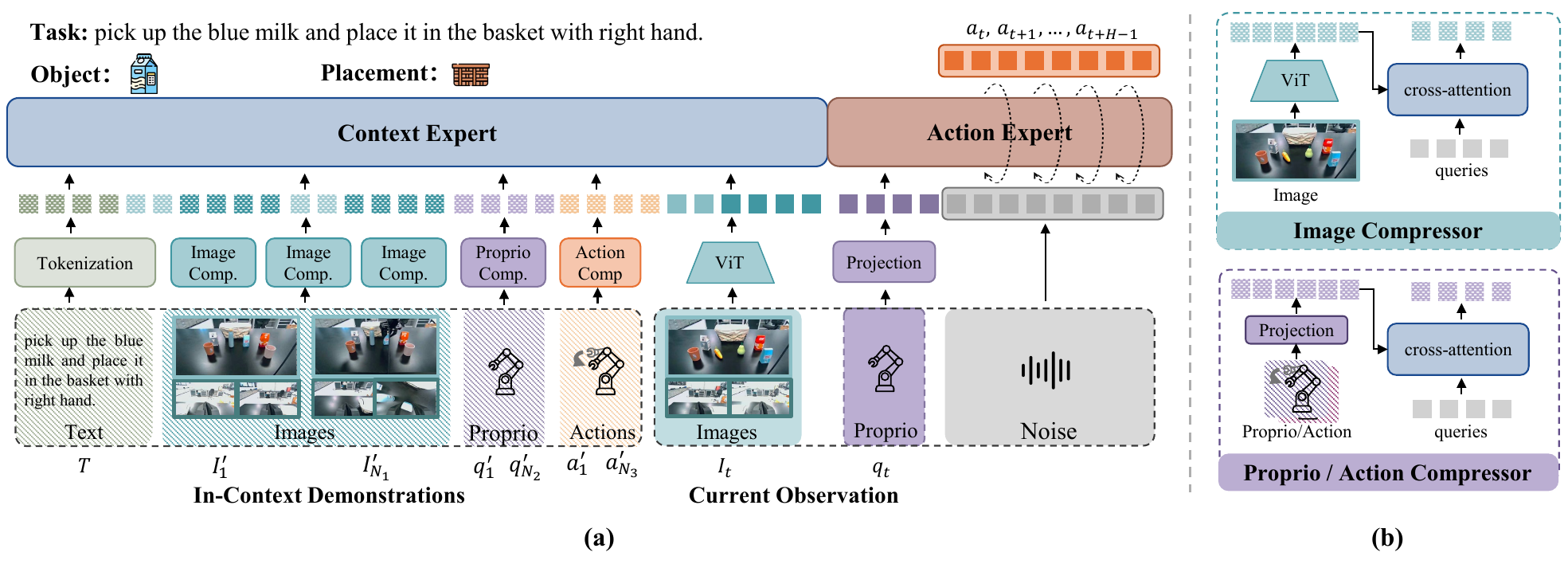}
    \caption{\textbf{ContextFlow Model Architecture.} We follow the mixture-of-experts design and employ two experts: the \textbf{context expert} and the \textbf{action expert}. The current observation, text prompt, and in-context demonstrations are first encoded and passed through \textit{multimodal context compressors}, and the resulting compressed tokens are routed to the context expert, while the current proprioceptive state and action noise tokens are routed to the action expert. The context expert encodes the current observation and contextual information, while the action expert performs flow matching steps to predict the actions. 
}
    \label{fig:network_arch}
\end{figure*}

\section{Method}
\label{method}

\subsection{Problem Formulation}
\noindent \textbf{Learning Objective.} We assume a distribution of tasks $\tau \sim \mathcal{T}$, where each task $\tau$ corresponds to a specific goal the robot must achieve, described by a textual instruction $T$ (e.g., \textit{``put the bowl on top of the cabinet''}). For each task $\tau$, demonstrations are drawn from $\mathcal{D}(d \mid \tau)$, where each demonstration is represented as $d = [T, X, Q, A]$: a task instruction $T$, an image sequence $X = (I^{'}_{1}, \dots, I^{'}_{T_d})$, proprio states (e.g., joint angles, gripper states) $Q = (q^{'}_{1}, \dots, q^{'}_{T_d})$, and actions $A = (a^{'}_{1}, \dots, a^{'}_{T_d})$. Each image $I^{'}_t \in \mathbb{R}^{H \times W \times 3}$, proprio state $q^{'}_t \in \mathbb{R}^{D_s}$, and action $a^{'}_t \in \mathbb{R}^{D_a}$ are recorded at step $t$ of a trajectory of length $T_d$. Our goal is to model the target data distribution: $\pi(\mathbf{a}_t \mid o_t, d)$, where we define $\mathbf{a}_t = a_{t:t+H-1}$ as the chunk of $H$ future actions. $o_t = (I_t, q_t)$ is the current observation and $d$ is the provided in-context demonstration. $I_t$ is the current image and $q_t$ is the current proprio state. $\pi(\mathbf{a}_t \mid o_t, d)$ can be learned through any proper \textit{conditional generative model}.

\noindent \textbf{Flow Matching Conditioned on Demonstration.}  While prior work~\cite{actiontoken} used an \textit{autoregressive model} to learn the distribution $\pi(\mathbf{a}_t \mid o_t, d)$, it requires tokenizing continuous actions into discrete tokens, which introduces quantization error and reduces accuracy. To effectively model the conditional distribution $\pi(\mathbf{a}_t \mid o_t, d)$, we leverage the Flow Matching (FM) framework, which learns a deterministic, time-dependent vector field that smoothly transports samples from a simple Gaussian prior to the target conditional distribution, allowing for direct modeling in \textit{continuous action spaces} without discretization. Given the current observation $o_t$ and the provided demonstration $d$, we construct a family of intermediate distributions that smoothly interpolate between a standard Gaussian prior at $s = 0$ and the target distribution $\pi(\mathbf{a}_t \mid o_t, d)$ at $s = 1$:
\begin{equation}
p_s(\mathbf{a}_t^s \mid \mathbf{a}_t) = \mathcal{N}(\mathbf{a}_t^s \mid s\, \mathbf{a}_t, (1-s)^2 I), \quad s \in [0,1].
\end{equation}

To sample intermediate actions, we draw noise $\epsilon \sim \mathcal{N}(0,I)$ and linearly interpolate: $\mathbf{a}_t^s = (1 - s)\, \epsilon + s\, \mathbf{a}_t$. For this linear interpolation path, the corresponding conditional velocity field can be computed in closed form~\cite{flowmatchtutorial}: $u(\mathbf{a}_t^s \mid \mathbf{a}_t) =  \mathbf{a}_t - \epsilon,$ representing the direct displacement from the noisy action chunk to the clean one. We parameterize our learnable vector field with a neural network: $u_\theta(\mathbf{a}_t^s, o_t, d, s),$ which conditions on the current noisy action chunk, observation, demonstration, and interpolation time $s$ (encoded with sinusoidal embeddings).

We train our model by minimizing the following Conditional Flow Matching loss:
\begin{equation}
\mathcal{L}_{\text{CFM}}(\theta)
= \mathbb{E}_{s,\,\mathbf{a}_t,\,\epsilon}
\left\|
u_\theta(\mathbf{a}_t^s, o_t, d, s)
- u(\mathbf{a}_t^s \mid \mathbf{a}_t)
\right\|^2,
\quad
\end{equation}
\text{where } $ s \sim \text{Beta}(\alpha, \beta),\;
\mathbf{a}_t \sim \pi(\mathbf{a}_t \mid o_t, d),\;
\epsilon \sim \mathcal{N}(0, I).$

During inference, we generate action sequences by integrating the learned velocity field from 0 to 1:
\begin{equation}
\mathbf{a}_t^{s+\delta} = \mathbf{a}_t^s + \delta\, u_\theta(\mathbf{a}_t^s, o_t, d, s), \quad \mathbf{a}_t^0 \sim \mathcal{N}(0,I),
\end{equation}
using a simple forward Euler solver with step size $\delta = \frac{1}{K}$ ($K = 10$ steps in our implementation). The resulting action chunk $\mathbf{a}_t^1$ provides the policy's next predicted actions.

\subsection{Model Architecture}

\noindent \textbf{Overview.}
To build the conditional flow matching model, we design the framework illustrated in Fig.~\ref{fig:network_arch}. We introduce a \textit{context expert} and \textit{multimodal context compressors} to efficiently encode in-context demonstrations. Following the mixture-of-experts design~\cite{shazeer2017outrageously,Glam,pi0}, the context expert encodes observation and compressed demonstration tokens, while the action expert processes proprio and action noise tokens, which interact with context tokens through attention. The action expert predicts a clean action chunk of horizon $H$ through $K$ flow-matching steps.

\noindent \textbf{Multimodal Context Compressor.}
Encoding long multimodal demonstrations is challenging due to the \textit{quadratic complexity of attention} over long sequences. Even tasks like ``put egg in box'' may produce demonstrations of length $T_d \approx 800$ steps, and multiple camera views introduce additional high-dimensional visual inputs $I^{'}_t \in \mathbb{R}^{H \times W \times 3}$ that further increase computational cost.
To reduce computational cost, we first perform temporal subsampling for each modality, resulting in shorter sequences of lengths \( N_1 \), \( N_2 \), and \( N_3 \) for images, states, and actions, respectively. To further reduce the redundancy, we adopt perceiver-style~\cite{perceiver,perceiverio} \textit{learnable context compressors} that condense multimodal demonstrations into compact latent representations while preserving task-relevant information. As shown in Fig.~\ref{fig:network_arch}, the learnable context compressors then take the input token sequences \(\mathbf{X} \in \mathbb{R}^{S \times D}\) from each modality and map them into fixed-size latent representations using alternating cross-attention and self-attention layers, where \(S\) denotes the input sequence length and \(D\) the embedding dimension.
A set of \(N_q\) learnable query tokens \(\mathbf{Z}_0 \in \mathbb{R}^{N_q \times D}\) forms the initial latent representation. The compressor then refines these latents through \(L\) layers that alternate between cross-attention and self-attention, each sub-layer using pre-normalization and a residual connection. The cross-attention layers let the latents \(\mathbf{Z}\) attend to the input tokens \(\mathbf{X}\) to absorb information from the demonstration, while the self-attention layers let the latents attend to one another and are followed by a token-wise feed-forward network. After the final layer, the compressed representation is the latent set \(\mathbf{Z}_{\text{out}} \in \mathbb{R}^{N_q \times D}\).

We design three separate compressors, \(\mathcal{C}_{\text{img}}\), \(\mathcal{C}_{\text{state}}\), and 
\(\mathcal{C}_{\text{action}}\), to encode image, state, and action sequences respectively into compact latent sets. For each camera view, the image compressor \(\mathcal{C}_{\text{img}}\) jointly compresses the visual tokens from all sampled demonstration frames into \(N_q\) latent tokens, with weights shared across camera views. The state and action compressors each produce \(N_q\) latent tokens. The outputs of all compressors are concatenated to form a unified context representation:
\begin{equation}
\mathbf{T}_{\text{context}} = 
[\mathbf{Z}_{\text{img}}; \mathbf{Z}_{\text{state}}; \mathbf{Z}_{\text{action}}] 
\in \mathbb{R}^{N_{\text{context}} \times D},
\label{eq:context_concat}
\end{equation}
where \(N_{\text{context}}\) is the total number of latent tokens after concatenation.

\begin{figure*}[!t]
\centering
\includegraphics[width=\linewidth]{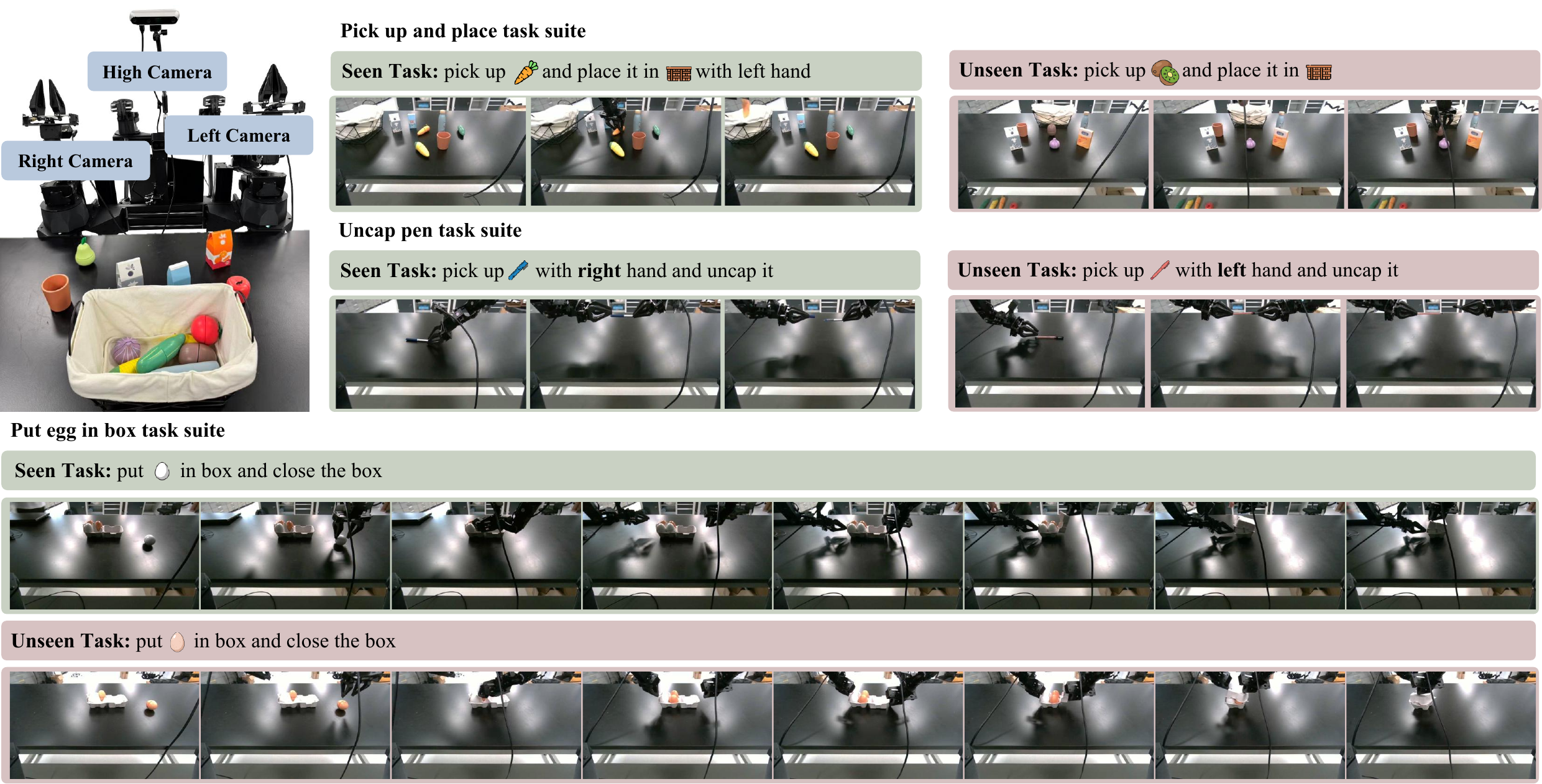}
    \caption{Examples of our real-world in-context experimental setup using the ALOHA robot platform. We evaluate three task suites. Each suite comprises multiple task configurations that differ in manipulated objects or the required left--right hand motion.}
    \label{fig:task}
\end{figure*}

The task instruction is tokenized using a text tokenizer~\cite{beyer2024paligemma}, and the current observation images are encoded with the same ViT used for demonstration frames. These tokens, together with the compressed demonstration context, are concatenated and passed to the context expert. During training, the context expert receives tokens from the current observation of a task (e.g., task $1$) along with randomly sampled demonstrations from the same task. At inference time, demonstrations from a novel task (e.g., task $2$) are supplied as in-context examples, allowing the model to perform previously unseen task configurations by leveraging the demonstration-driven context without any fine-tuning.

\noindent \textbf{Context-Conditioned Flow Matching.} 
The current proprio state of the robot arm and the noise action tokens are routed to the action expert, where they attend to the context expert tokens. This attention mechanism allows the action expert to aggregate relevant information from both the in-context demonstration and the current observation, enabling action prediction on unseen task configurations. The noise action tokens are then updated through $K$ flow-matching steps, each involving attention interactions between the noise action tokens and the context expert tokens. To enhance the efficiency of flow matching, we implement a block-wise causal attention mechanism that enforces a causal dependency from context to action tokens: action tokens can attend to context tokens, but context tokens cannot attend to action tokens. This causal design allows the caching of context tokens during inference, eliminating the need to recompute context tokens at each flow matching step.

\section{Experimental Setup}
\label{experiments}
\subsection{Simulation Benchmark Setup} 
\noindent \textbf{Simulation Benchmark.} We compare ContextFlow with the baselines on the LIBERO benchmark~\cite{libero}, a simulated environment featuring a Franka Emika Panda robotic arm. The benchmark includes multimodal demonstrations consisting of \textit{RGB images} from the workspace and wrist cameras, \textit{robot proprioception}, and \textit{text task instructions}. Our evaluation focuses on two task suites: spatial reasoning (LIBERO-Spatial) and object manipulation (LIBERO-Object). To increase the diversity of training data, we additionally include demonstrations from LIBERO-Goal and LIBERO-10 during training.

\noindent \textbf{Training and Testing Split.} Each suite comprises 10 unique tasks with a total of 500 expert demonstrations per suite.
We divide the $10$ tasks of each evaluated task suite into $8$ \textit{seen tasks} and $2$ \textit{unseen task configurations}. This split is randomly predefined in the benchmark setup and consistent across all experiments. Seen tasks are used during training, while unseen task configurations are reserved for inference. We focus our evaluation on \textit{unseen configurations within related primitives}, which reflect the model's generalization ability. We report the average success rate over $50$ inference trials per task.

\subsection{Real-World Experiment Setup}
\label{sec:real_world_setup}
We conducted real-world experiments on an ALOHA robot platform, as shown in Fig.~\ref{fig:task}. The setup includes one high camera and two wrist cameras, and demonstrations were collected through human teleoperation. We evaluated on three task suites: (1) ``pick up and place,'' following the instruction template \texttt{``Pick up the \textless object\textgreater{} and place it in the basket with \textless left/right\textgreater{} hand.''} (2) ``uncap pen,'' following the instruction template \texttt{``Pick up \textless type of pen\textgreater{} with \textless left/right\textgreater{}, grasp the cap with another hand and uncap it.''} (3) ``put egg in box,'' following the instruction template \texttt{``Pick up \textless object\textgreater{} with right hand and place it in box and close the box.''} For each task suite, there are different \textit{task configurations} with different \textit{objects} or \textit{left--right hand motion trajectories}. In total, these three task suites comprise 21 task configurations and 1{,}064 training demonstrations. The average episode lengths are 234, 534, and 736 timesteps, respectively. Notably, the latter two suites involve \textit{long-horizon}, fine-grained bimanual manipulation requiring precise closed-loop feedback. We also included 4 extra bimanual tasks (handover, cup stack, stir, and water wipe) with 254 demonstrations for training only.
Examples of seen and unseen task configurations are shown in Fig.~\ref{fig:task}. Detailed task names and stats are provided in the \textit{Supplementary Material}.

\subsection{Implementation Details}
We use SigLIP~\cite{siglip} as the vision encoder, processing observation images at a resolution of $224 \times 224$. Observation images at each timestep consist of all camera views. The transformer architectures of the context expert and the action expert follow a structure similar to Gemma~\cite{gemma}, but are significantly smaller. By default, the model weights for both SigLIP and the action expert are initialized from $\pi_0$~\cite{pi0}. The weights of the context expert are randomly initialized. We use LoRA~\cite{hu2022lora} to fine-tune the action expert, with a rank of 32 and an $\alpha$ of 32. The vision encoder is fully fine-tuned, and the context expert is trained in full. We train the models for 20K iterations with a batch size of 32. We use a cosine learning rate schedule with a peak of $2.5 \times 10^{-5}$ and a minimum of $2.5 \times 10^{-6}$, including a warm-up phase over the first 1,000 steps. For in-context demonstrations, we sample $N_1 = 8$ image frames and $N_2 = 128$ proprioceptive states and $N_3 = 128$ actions. Each compressor uses 32 learnable query tokens per modality and is refined through stacked cross-attention and self-attention layers: the image compressor has 4 layers, and the proprioceptive state and action compressors each have 2 layers. For flow matching, we set the Beta distribution parameters to $\alpha = 1.5$ and $\beta = 1$.

\begin{table}[!t]
\centering
\caption{\textbf{Comparison of models on LIBERO simulation.} 
Models are trained on seen tasks and evaluated on unseen task configurations.  
The two unseen task configurations on LIBERO-Spatial are ``pick up the black bowl on the cookie box and place it on the plate'' and ``pick up the black bowl next to the plate and place it on the plate.'' 
The two unseen task configurations on LIBERO-Object task suite are ``pick up the milk and place it in the basket'' and ``pick up the tomato sauce and place it in the basket.'' 
FT means fine-tuning on unseen task configurations with 1 demonstration. 
ContextAR is our improved in-context autoregressive baseline for fair comparison with ContextFlow.}

\resizebox{0.99\linewidth}{!}{
\begin{tabular}{l ccc ccc c}
\toprule
\multirow{3}{*}{\textbf{Model}} 
& \multicolumn{3}{c}{\textbf{LIBERO-Spatial}} 
& \multicolumn{3}{c}{\textbf{LIBERO-Object}} 
& \multirow{2}{*}{\textbf{Avg.}} \\
\cmidrule{2-4} \cmidrule{5-7}
& \textbf{S$_1$} (\includegraphics[height=0.7em]{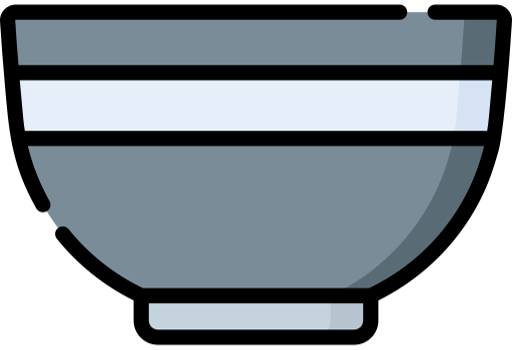} on 
\includegraphics[height=0.7em]{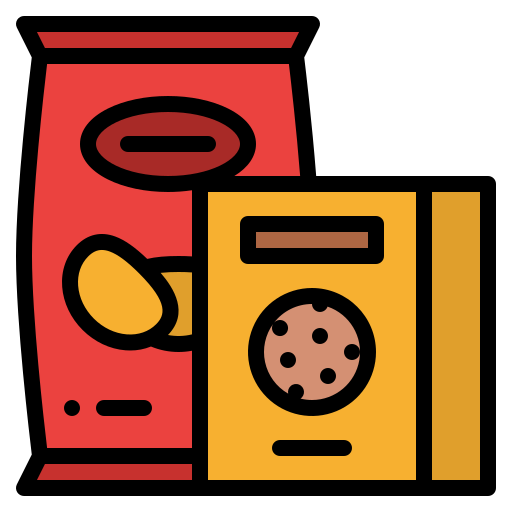})
& \textbf{S$_2$} (\includegraphics[height=0.7em]{figures/Fig_Icon_Bowl.png} \small{next to} 
\includegraphics[height=0.7em]{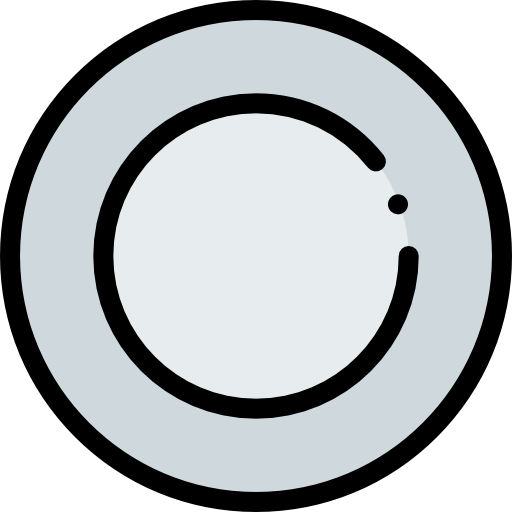})
& \textbf{Avg.}
& \textbf{O$_1$} (\includegraphics[height=0.7em]{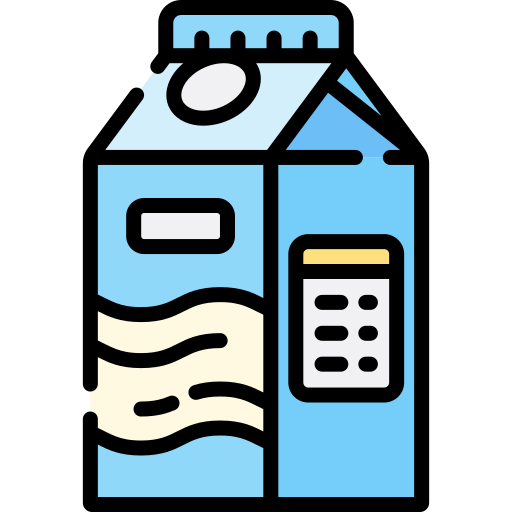} in 
\includegraphics[height=0.7em]{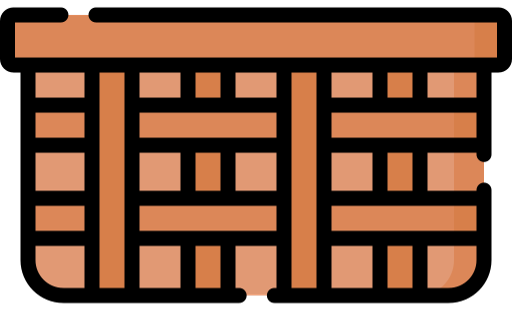})
& \textbf{O$_2$} (\includegraphics[height=0.7em]{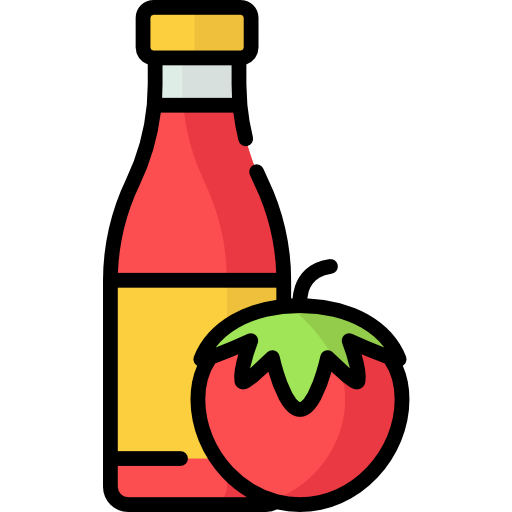} in 
\includegraphics[height=0.7em]{figures/Fig_Icon_Basket.png})
& \textbf{Avg.}
& \\
\midrule
OpenVLA-OFT~\cite{kim2025fine} & \textbf{94.0} & 32.0 & 63.0 & 0.0 & 34.0 & 17.0 & 40.0 \\
$\pi_0$~\cite{pi0} & 52.0 & 0.0 & 26.0 & 46.0 & 80.0 & 63.0 & 44.5\\
\textcolor{mediumgray}{$\pi_0$~\cite{pi0} FT} & \textcolor{mediumgray}{94.0} & \textcolor{mediumgray}{32.0} & \textcolor{mediumgray}{63.0} & \textcolor{mediumgray}{80.0} & \textcolor{mediumgray}{84.0} & \textcolor{mediumgray}{82.0} & \textcolor{mediumgray}{72.5}\\
\midrule
ICRT (AR baseline)~\cite{icrt} & 54.0 & 0.0 & 27.0 & 4.0 & \textbf{96.0} & 50.0 & 38.5\\
ContextAR (AR baseline$^{+}$)  & \textbf{94.0} & 12.0 & 53.0 & 16.0 & 92.0 & 54.0 & 53.5 \\
ContextFlow-Plain (Ours) & 90.0 & 20.0 & 55.0 & \textbf{92.0} & 54.0 & 73.0 & 64.0\\
ContextFlow (Ours) & 86.0 & \textbf{42.0} & \textbf{64.0} & 76.0 & 90.0 & \textbf{83.0} & \textbf{73.5}\\
\bottomrule 
\end{tabular}
}
\label{tab:model_comparison}
\end{table}

\subsection{Baselines}
\noindent\textbf{Autoregressive-Based In-context Imitation Learning Methods.} We compare with autoregressive baselines to show the advantage of flow matching in the in-context imitation learning task setting. We compare with the ICRT~\cite{icrt}, which is a recently proposed in-context imitation learning model that formulates the learning process as a next-token prediction task. Images and proprioceptive states are encoded using attention pooling to obtain compact state representations. Demonstrations are structured as sequences of states and actions. A causal Transformer (\ie, LLaMA2~\cite{touvron2023llama}) takes as input the current state, the history of past states and actions, and demonstration sequences to predict the next action within a specified horizon. Following the original ICRT implementation, we use LLaMA2-Base as the backbone LLM. During inference, an episode from the corresponding task will be provided as the prompt. ICRT uses modalities of images, and state/actions as in-context demonstration. In addition to ICRT, we implemented an autoregressive baseline, the \textit{in-context autoregressive model (ContextAR)}, which uses the same vision backbone as ContextFlow and is initialized from $\pi_0$ to ensure a fair comparison between autoregressive and flow-matching approaches. ContextAR accepts all modalities: images, text, and states/actions as context.
    
\noindent\textbf{Large Scale Vision-Language-Action (VLA) Models.} These VLA models are pre-trained on large-scale robot data and fine-tuned on seen tasks of LIBERO data. We provide text instructions of unseen task configurations to these VLA models to prompt them to predict actions for unseen task configurations. We compared with $\pi_0$ and OpenVLA-OFT~\cite{kim2025fine}. 
    
\noindent\textbf{VLA Fine-tuned with 1-shot Demonstration from Unseen Task.} VLA models cannot accept in-context demonstrations as input. Therefore, we further fine-tuned $\pi_0$ with one demonstration from an unseen task. Although task-specific fine-tuning methods are not as flexible as in-context learning methods that do not require fine-tuning, they can serve as a reference.

\section{Results}
\subsection{Comparison with Baselines}
\noindent\textbf{Autoregressive vs.\ Flow-Matching Models.}
As shown in Tab.~\ref{tab:model_comparison}, we compare flow-matching policies against autoregressive (AR) baselines in a controlled manner. We treat ICRT~\cite{icrt} as a representative autoregressive baseline, and introduce ContextAR as a stronger autoregressive baseline that incorporates $\pi_0$~\cite{pertsch2025fast,pi0} pre-training. To isolate the effect of the modeling paradigm (autoregression vs.\ flow matching) from other architectural choices, our key comparison is between ContextFlow-Plain (flow matching, without context compressors) and ContextAR.  Both models use the same vision backbone and are initialized from $\pi_0$, making this our most controlled comparison between autoregressive and flow-matching action modeling. Under this setup, ContextFlow-Plain outperforms ContextAR by 10.5 percentage points in success rate, providing evidence that flow matching is beneficial for in-context imitation learning. ContextFlow also achieves higher throughput than AR baselines, as shown in the Supplementary Material.

\noindent \textbf{Text vs.\ In-Context Demonstrations.}
We compared in-context imitation learning models with VLA models. The results show that in-context imitation learning models are usually better than VLA models. ContextAR outperforms OpenVLA-OFT~\cite{kim2025fine} by 13.5 percentage points (see Tab.~\ref{tab:model_comparison}), showing that a text prompt alone is insufficient for the generalization to unseen tasks. Providing detailed in-context demonstration (sequence of images, states, and actions) benefits the generalization to unseen task configurations.

\noindent \textbf{Parametric Fine-tuning vs.\ In-Context Learning.}
In our in-context learning setting, methods receive a single demonstration from each unseen task at test time. To compare against parametric adaptation, we additionally fine-tune the VLA model $\pi_0$ with one demonstration from each unseen task at test time, denoted $\pi_0$ (FT). Our proposed ContextFlow performs comparably to $\pi_0$ (FT) in success rate (73.5\% vs.\ 72.5\%; see Tab.~\ref{tab:model_comparison}). Note that the in-context learning paradigm does not require any fine-tuning on demonstrations from unseen task configurations, enabling much faster adaptation. Moreover, aside from the additional computation cost, task-specific fine-tuning can also lead to catastrophic forgetting of previously learned tasks. These results highlight the advantage of our in-context learning model over baselines with task-specific fine-tuning.

\subsection{Ablations}
\noindent \textbf{The Effectiveness of Context Compressor.}
As shown in Tab.~\ref{tab:model_comparison}, adding context compressors further boosts performance, with ContextFlow outperforming ContextFlow-Plain by 9.5 and ContextAR by 20 percentage points in average success rate. We conduct an ablation to examine the contribution of the proposed context compressors for different modalities in the in-context demonstrations. As shown in Tab.~\ref{tab:ablation_compressor}, enabling only the image compressor improves the average success rate by 4 percentage points. Activating both compressors improves overall performance over ContextFlow-Plain, with the largest gains in spatial reasoning and average success rate, indicating that jointly compressing visual and state/action contexts is important for effective in-context flow matching. 

\noindent \textbf{Influence of Context Modalities.}
The text prompting and in-context demonstrations include three modalities: (1) text instruction, (2) images, and (3) proprioceptive states and actions. To assess the contribution of each modality to in-context imitation learning, we conducted ablation experiments by removing one modality at a time from our full model, as shown in Table~\ref{tab:ablation_modality}. The results show that all three modalities are important for both seen and unseen task configurations, with a greater impact observed on unseen performance. Removing the text modality leads to a 9.0 points drop in performance on average for unseen task configurations. Among all modalities, proprioceptive states and actions are the most critical: removing them causes a 44.5 points drop on average for unseen task configurations.

\noindent \textbf{Hyperparameter Analysis.}
We further study how the length of in-context demonstrations influences model performance in Tab.~\ref{tab:ablation_num_frames}. 
When we change the number of frames from 4 to 8 while keeping the number of states/actions the same, there is an improvement by 3.0 points in success rate on average, suggesting more visual information would help in-context imitation learning. We then fix the number of frames, and vary the number of states/actions, and find that 128 is the optimal number for states/actions in context.

\noindent \textbf{Comparison with Other Action Heads.}
To further study the design choices within continuous parallel action heads, we compare MLP L1 regression, diffusion, and flow matching. The success rates on LIBERO are reported in Tab.~\ref{tab:compressor_action_decoder_ablation}. MLP L1 regression achieves a success rate of 58\%, outperforming ContextAR by 4.5 points, which suggests that continuous parallel prediction is effective even with a simple regression head. The diffusion head converges slowly in our setting, achieving 16.5\% success after 20k iterations and 33.5\% after 100k iterations. Among the continuous parallel action heads, flow matching achieves the best overall performance with a success rate of 73.5\%. We therefore adopt flow matching as the default action head, which is also aligned with the $\pi_0$ pre-training objective.

\begin{table}[t]
\centering

\begingroup
\setlength{\tabcolsep}{0pt}
\renewcommand{\arraystretch}{1.12}

\begin{tabular*}{\linewidth}{@{}p{0.48\linewidth}@{\extracolsep{\fill}}p{0.48\linewidth}@{}}

\multicolumn{2}{@{}c@{}}{%
\begin{minipage}[t]{0.72\linewidth}
\centering
\caption{\textbf{Ablation study on context compressors.}}

{\scriptsize
\setlength{\tabcolsep}{1.8pt}
\renewcommand{\arraystretch}{0.95}
\begin{tabular}{ccc|ccc}
\toprule
\textbf{Model} & \textbf{$\mathcal{C}_{\text{img}}$} & \textbf{$\mathcal{C}_{\text{state}}$ \& $\mathcal{C}_{\text{action}}$} & \textbf{Spatial} & \textbf{Object} & \textbf{Avg.} \\
\midrule
ContextFlow-Plain & \xmark & \xmark & 55.0 & 73.0 & 64.0 \\
& \cmark & \xmark & 51.0 & \textbf{85.0} & 68.0 \\
ContextFlow & \cmark & \cmark & \textbf{64.0} & 83.0 & \textbf{73.5} \\
\bottomrule
\end{tabular}
}
\label{tab:ablation_compressor}
\end{minipage}%
}
\\

\begin{minipage}[t]{\linewidth}
\centering
\caption{\textbf{Ablation study on different modalities in the demonstrations.} * indicates the default model.}

{\scriptsize
\setlength{\tabcolsep}{1.8pt}
\renewcommand{\arraystretch}{0.95}
\begin{tabular}{@{}ccccccc@{}}
\toprule
& \textbf{State} & \textbf{Image} & \textbf{Text}
& \textbf{Spatial} & \textbf{Object} & \textbf{Avg.} \\
\midrule
& \xmark & \cmark & \cmark & 30.0  & 28.0 & 29.0 \\
& \cmark & \xmark & \cmark & 42.0 & 48.0 & 45.0 \\
& \cmark & \cmark & \xmark & 59.0 & 70.0 & 64.5 \\
$*$ & \cmark & \cmark & \cmark & \textbf{64.0} & \textbf{83.0} & \textbf{73.5} \\
\bottomrule
\end{tabular}
}
\label{tab:ablation_modality}
\end{minipage}
&
\begin{minipage}[t]{\linewidth}
\centering
\caption{\textbf{Ablation study on the number of images and states in in-context demonstrations.}}

{\scriptsize
\setlength{\tabcolsep}{1.8pt}
\renewcommand{\arraystretch}{0.95}
\begin{tabular}{@{}cccccc@{}}
\toprule
\multicolumn{2}{c}{\textbf{\# Images}} & \textbf{\# States}
& \textbf{Spatial} & \textbf{Object} & \textbf{Avg.} \\
\midrule
& 4 & 128 & 62.0 & 79.0 & 70.5 \\
& 8 & 64  & 53.0 & \textbf{87.0} & 70.0 \\
$*$ & 8 & 128 & \textbf{64.0} & 83.0 & \textbf{73.5} \\
& 8 & 256 & 52.0 & 81.0 & 66.5 \\
\bottomrule
\end{tabular}
}
\label{tab:ablation_num_frames}
\end{minipage}
\\

\begin{minipage}[t]{\linewidth}
\centering
\caption{\textbf{Ablation of action heads.}}

{\scriptsize
\setlength{\tabcolsep}{1.8pt}
\renewcommand{\arraystretch}{0.95}
\begin{tabular}{@{}lcccc@{}}
\toprule
\textbf{Action Head} & \textbf{Iterations} & \textbf{Spatial} & \textbf{Object} & \textbf{Avg.} \\
\midrule
MLP L1 & 20k & 26.0 & \textbf{90.0} & 58.0 \\
Diffusion & 100k & 30.0 & 37.0 & 33.5 \\
Flow Matching & 20k & \textbf{64.0} & 83.0 & \textbf{73.5} \\
\bottomrule
\end{tabular}
}
\label{tab:compressor_action_decoder_ablation}
\end{minipage}
&
\begin{minipage}[t]{\linewidth}
\centering
\caption{\textbf{Cumulative MSE.}}

{\scriptsize
\setlength{\tabcolsep}{1.8pt}
\renewcommand{\arraystretch}{0.95}
\begin{tabular}{@{}lc@{}}
\toprule
\textbf{Model} & \textbf{MSE ($\mathrm{m}^2$)} \\
\midrule
ContextAR (baseline$^{+}$) & 4.7 \\
ContextFlow-Plain & 3.0 \\
ContextFlow & 2.2 \\
\bottomrule
\end{tabular}
}
\label{tab:cumulated_error}
\end{minipage}

\end{tabular*}

\endgroup
\end{table}

\begin{table}[t]
\centering
\caption{\textbf{Results with Expanded Training Data and Evaluation Tasks
}}
\label{tab:libero90_train}
\scriptsize
\begin{tabular}{lccccc}
\toprule
\textbf{Training Data} & \textbf{Spatial} & \textbf{Object} & \textbf{Goal} & \textbf{10} & \textbf{Avg} \\
\midrule
4 suites & \textbf{64.0} & 83.0 & 0.0 & 0.0 & 36.8 \\
4 suites + LIBERO-90 & 55.0 & \textbf{85.0} & \textbf{29.0} & \textbf{3.0} & \textbf{43.0} \\
\bottomrule
\end{tabular}
\end{table}

\noindent \textbf{Results on Expanded Training Data and Evaluation Tasks.}
In the previous models, we demonstrated generalization only on LIBERO-Spatial and LIBERO-Object. We investigate whether generalization on these more challenging task suites can be improved by increasing the amount of training data. To this end, we incorporate LIBERO-90 into ContextFlow's training set and evaluate on additional unseen tasks from LIBERO-Goal and LIBERO-10, with two unseen tasks from each suite. The results are reported in Tab.~\ref{tab:libero90_train}. With this more diverse training data, ContextFlow achieves improved results on the previously unsolved LIBERO-Goal and LIBERO-10 tasks, raising the overall average from 36.8 to 43.0 points.

\subsection{Results on Real-World Experiments}
We compared the proposed ContextFlow with ICRT~\cite{icrt} in real-world experiments in Tab.~\ref{tab:real_world}. Each model was evaluated for 10 trials per task. ContextFlow achieved reasonable performance on both single-arm pick-and-place configurations and fine-grained bimanual configurations. In contrast, ICRT succeeded in only 2/10 trials on the pear pick-and-place task and struggled with the fine-grained bimanual tasks, which require high-frequency control and closed-loop feedback. These results suggest that ContextFlow can reliably leverage in-context demonstrations to solve unseen real-world manipulation task configurations. More real-world experiments and rollout videos are provided in the \textit{Supplementary Materials}.

\begin{table}[t]
\centering
\caption{\textbf{Performance on unseen task configurations on a real-world robot.} See Sec.~\ref{sec:real_world_setup} for task instruction templates. Elements in different task configurations are listed in the table.}
\label{tab:real_world}

{\scriptsize
\setlength{\tabcolsep}{3pt}

\begin{tabular*}{0.98\linewidth}{@{\extracolsep{\fill}}lcccccc}
\toprule
\multirow{3}{*}{\textbf{Model}} &
\multicolumn{4}{c}{\textbf{Single-arm Pick-and-place tasks}} &
\multicolumn{2}{c}{\textbf{Fine-grained bimanual tasks}} \\

\cmidrule{2-5} \cmidrule{6-7}

& \multicolumn{2}{c}{\textbf{Left hand}} &
\multicolumn{2}{c}{\textbf{Right hand}} &
\textbf{Uncap (Left Hand)} & \textbf{Put in box} \\

& \includegraphics[height=0.9em]{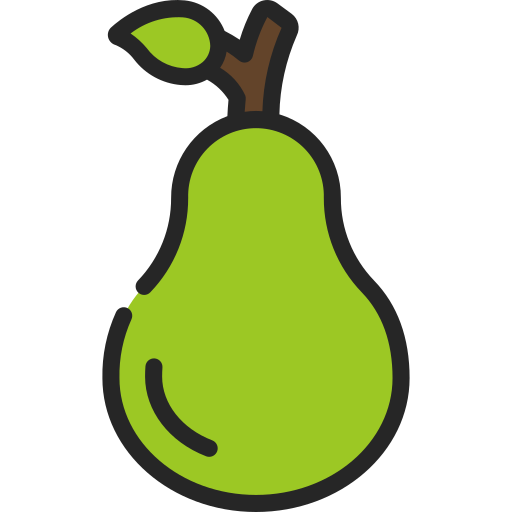} Pear
& \includegraphics[height=0.9em]{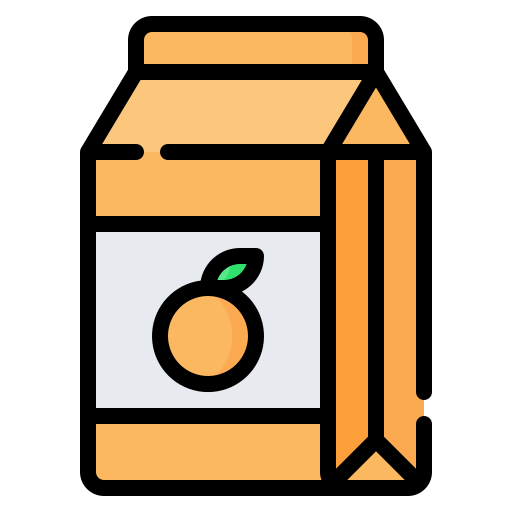} Orange
& \includegraphics[height=0.9em]{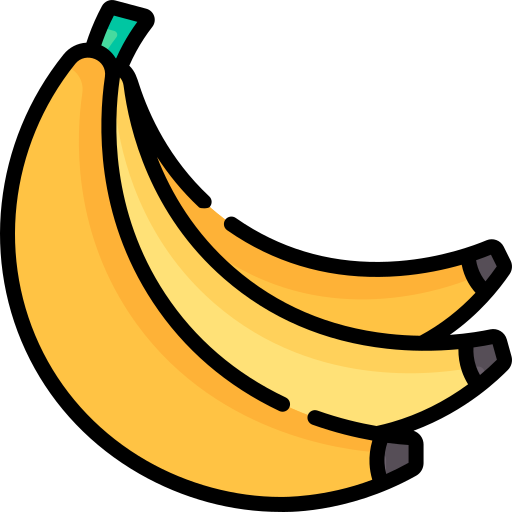} Banana
& \includegraphics[height=0.9em]{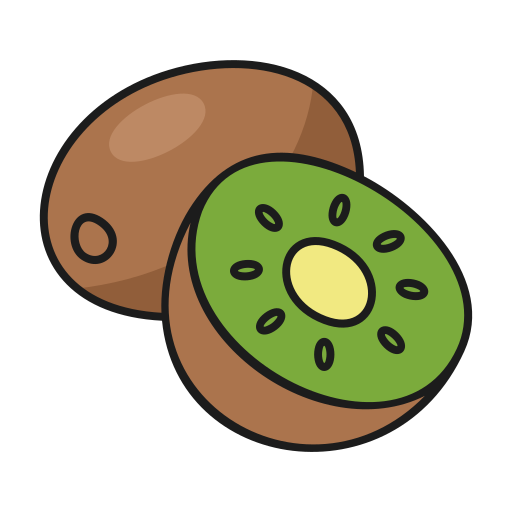} Kiwi
& \includegraphics[height=0.9em]{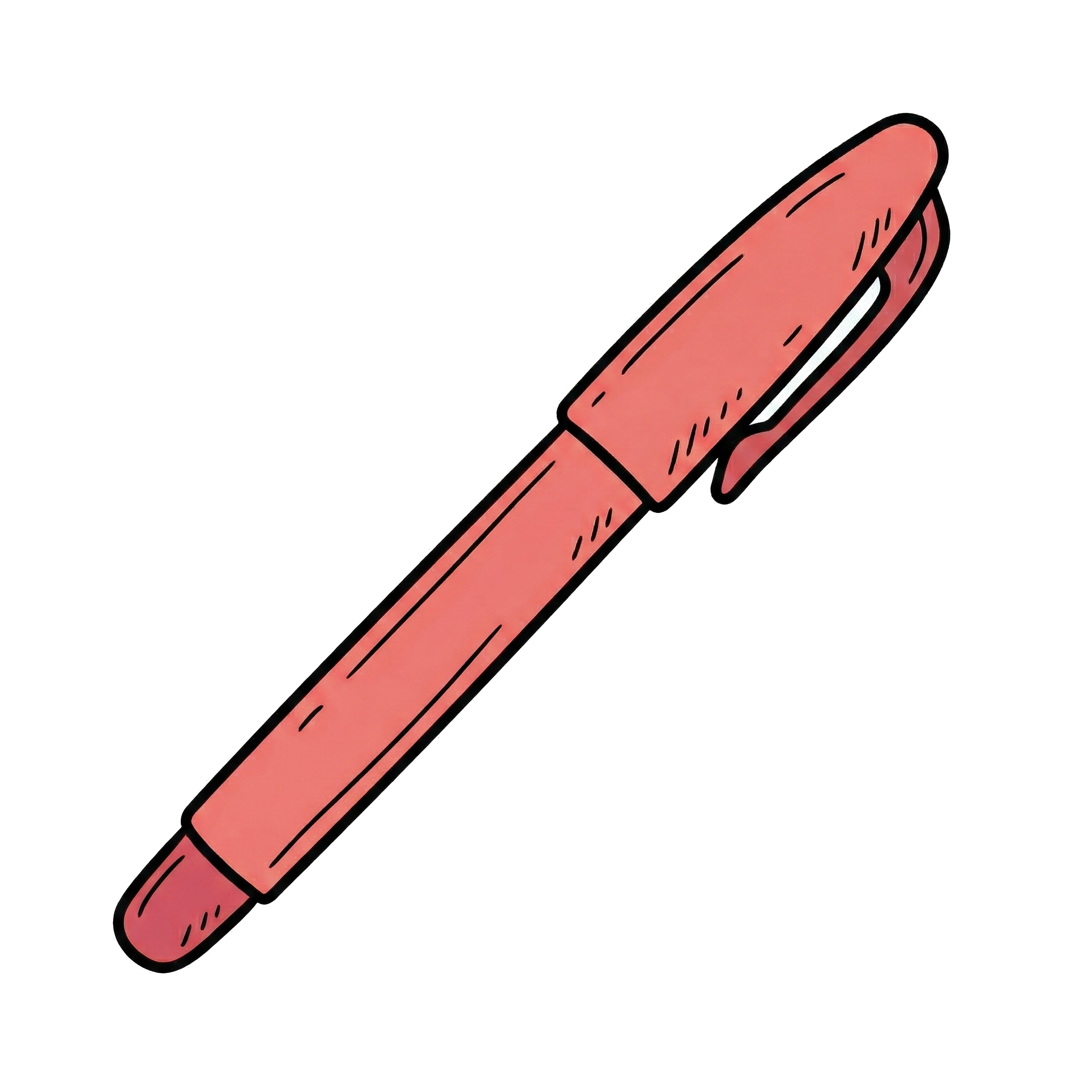} Red Pen
& \includegraphics[height=0.9em]{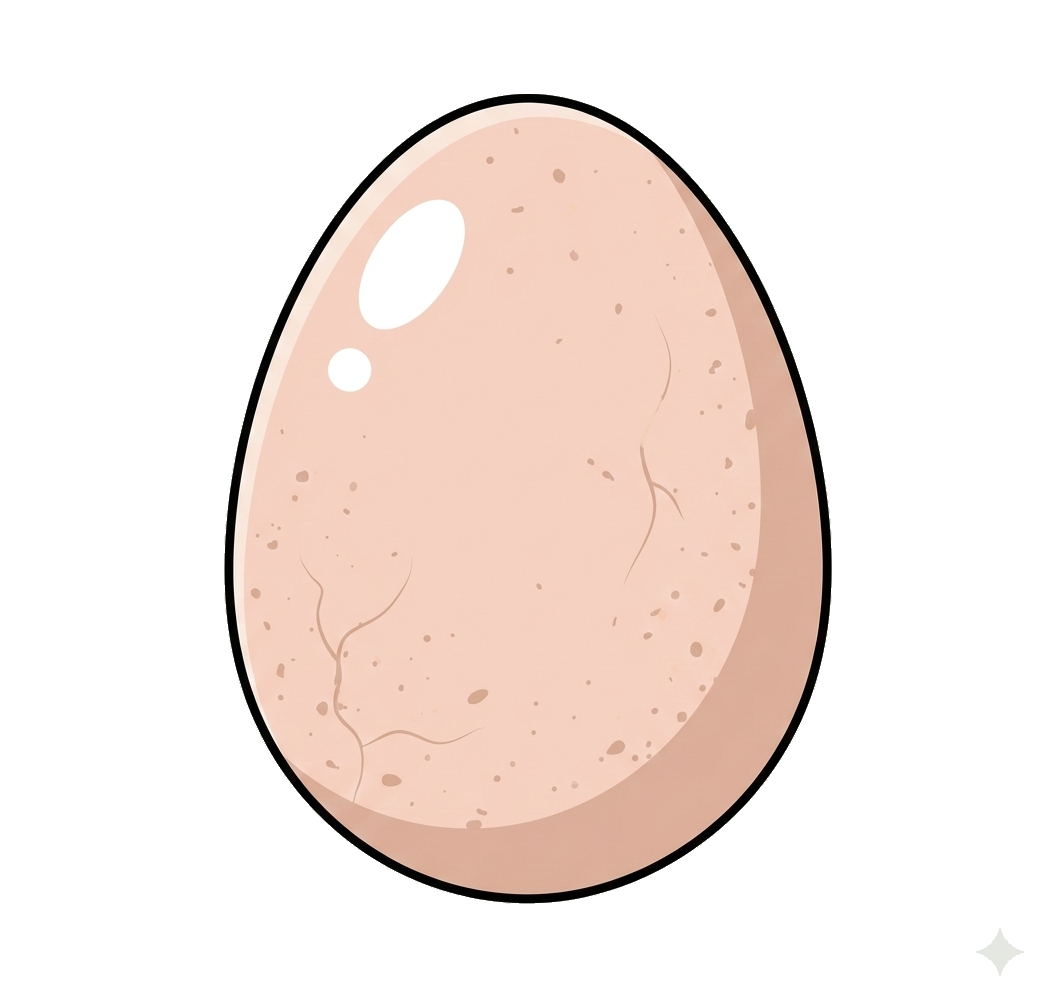} Red Egg \\

\midrule

ICRT~\cite{icrt} & \textbf{2/10} & 0/10 & 0/10 & 0/10 & 0/10 & 0/10 \\
ContextFlow      & \textbf{2/10} & \textbf{5/10} & \textbf{4/10} & \textbf{4/10} & \textbf{4/10} & \textbf{6/10} \\

\bottomrule
\end{tabular*}
}
\end{table}

\begin{figure}[t]
\centering
\newlength{\FigH}
\setlength{\FigH}{3.2cm}

\begin{minipage}[t]{0.52\linewidth}
\centering
\includegraphics[height=\FigH]{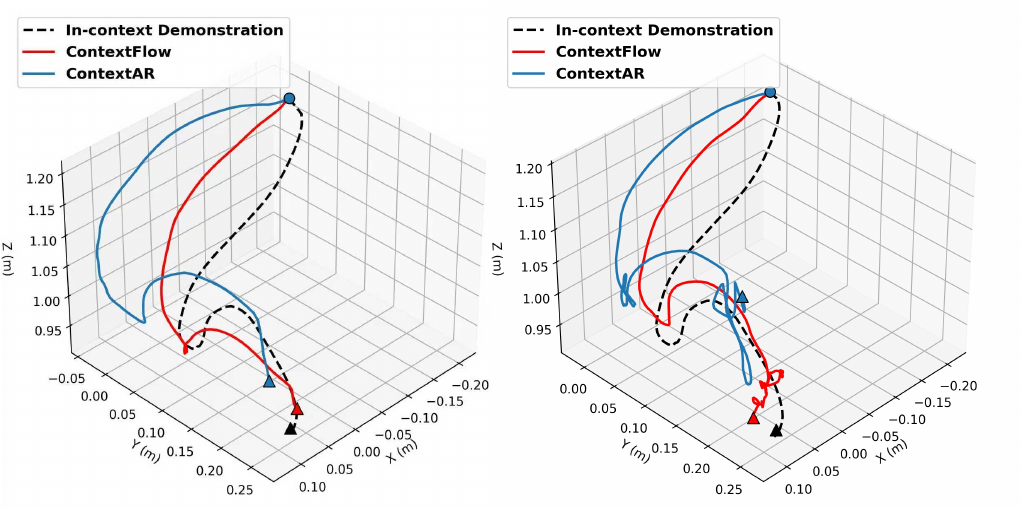}

\captionof{figure}{\textbf{Visualization of in-context demonstration, and success, failure of ContextFlow and ContextAR.}
Left: Success trajectories.
Right: Failure trajectories.}\label{fig:analyses}
\end{minipage}
\hfill
\begin{minipage}[t]{0.43\linewidth}
\centering
\includegraphics[height=\FigH]{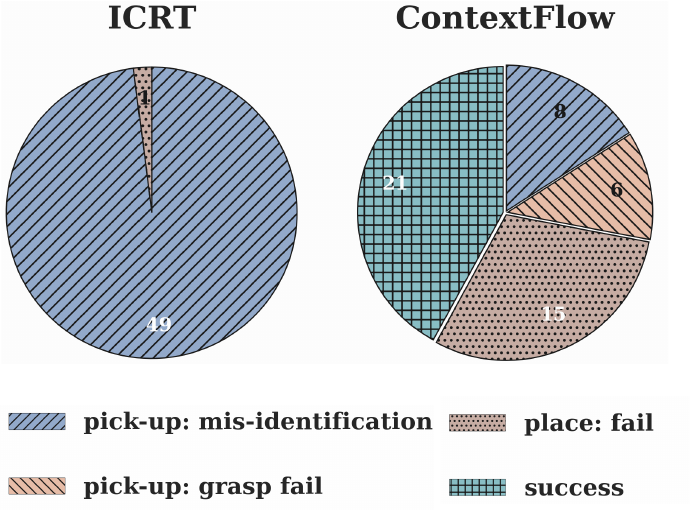}

\captionof{figure}{\textbf{Failure mode analysis.}
We analyze rollouts of ICRT and ContextFlow and categorize failure modes into three categories.}\label{fig:failure_mode}
\end{minipage}

\end{figure}

\section{Analysis}

\noindent \textbf{Compounding Error.} To quantify compounding error, we compute the mean squared error (MSE) between the evaluation rollout and the reference expert trajectory over time. We visualize the error on a two-stage pick-and-place task ($S_{2}$ in Tab.~\ref{tab:model_comparison}) using 50 test episodes, plotting the mean and the 20--80 percentile band in Fig.~\ref{fig:overview}~(d). We observe a mild decrease in MSE starting around step 60, aligning with the transition that marks the end of stage 1 (approximately steps 50--65). ContextFlow and ContextAR exhibit similar errors during the first 30 steps; however, beyond this point, ContextAR's error grows noticeably over time. In this evaluated task, this suggests that autoregressive models tend to accumulate larger compounding errors than flow-matching models in robot manipulation. The distribution shift between seen and unseen configurations further exacerbates this compounding effect and leads to lower success rates on unseen configurations. We also visualize the expert in-context demonstration trajectory and the rollout end-effector trajectories in 3D space. Further, we sum the per-step errors over the trajectory as a metric. From Tab.~\ref{tab:cumulated_error}, we can see that both flow matching reduces MSE accumulation by 1.7, and compression can further reduce MSE by 0.8. Successful rollout trajectories are shown in Fig.~\ref{fig:analyses} left. Although both ContextFlow and ContextAR succeed, the ContextFlow trajectory aligns more closely with the expert demonstration. Failed trajectories are shown in Fig.~\ref{fig:analyses} right, where we observe clear detours in the pick-up and place stages, respectively.

\begin{figure}[t]
\centering
\includegraphics[width=0.98\linewidth]{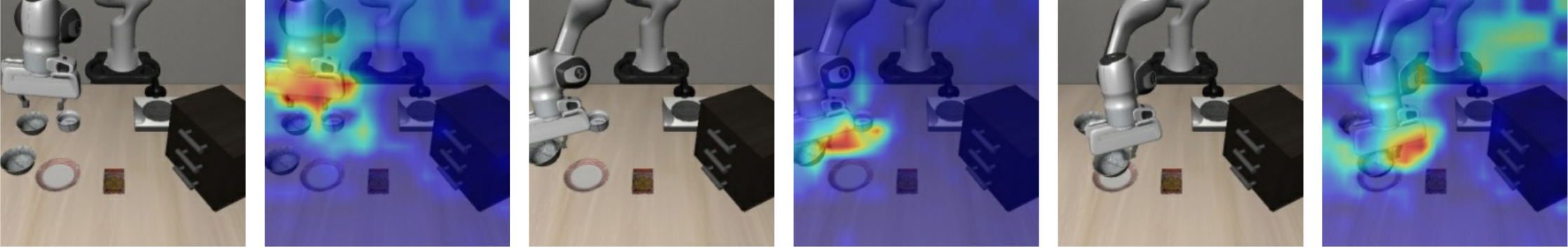}
    \caption{\textbf{Context compressor selects task-relevant, compact information.} Visualization of image-compressor attention on selected frames (3/6/8), showing that the attended image regions vary across time. In-context demonstration frames and their corresponding attention maps are shown interleaved. }
    \label{fig:attention}
\end{figure}

\noindent \textbf{Failure Mode.} We group failure modes into three categories at different stages: \textit{pick-up: misidentification}, \textit{pick-up: grasp fail}, and \textit{place: fail}. As shown in Fig.~\ref{fig:failure_mode}, ICRT frequently fails in the first stage by reaching a location near, but not on, the target object. This may stem from its state-token design: the visual latent and proprioceptive latent are attention-pooled into a single token, potentially discarding spatial detail needed for accurate object identification. For ContextFlow, the failure cases are more evenly distributed across the three stages with more examples on the last one. We provide visualization of different failure modes and analysis in \textit{Supplementary Materials}.

\noindent \textbf{Attention in Image Compressor.}
To understand what the image compressor learns, we visualize the attention of its learnable queries over frames in the in-context demonstration. For visualization, we average the attention weights over all 32 queries. The subsampled in-context demonstration has length 8, and we show selected frames (3/6/8) in Fig.~\ref{fig:attention}. 
We observe that the learnable queries attend to task-relevant regions according to task progress. Across time, the attention pattern changes with the task phase: before grasping, it concentrates on the gripper; after the object is grasped, it shifts to the object; and during placing, it focuses on the plate at the target location.

\section{Conclusion}

In this paper, we propose ContextFlow, an in-context imitation learning framework for robotics based on conditional flow matching. Unlike autoregressive models that predict discrete action tokens sequentially, ContextFlow predicts continuous actions in parallel conditioned on both the current observation and in-context demonstrations. Our empirical results suggest that flow matching appears beneficial for generalizing to unseen task configurations, where continuous parallel action prediction helps alleviate compounding error. We further find that multimodal context compressors reduce spatial and temporal redundancy in demonstrations and further improve performance. However, the generalization demonstrated in this work is mainly limited to unseen objects and spatial arrangements within tasks that share similar underlying primitives, rather than to fundamentally different task types. Scaling up the diversity and quantity of training data may further improve generalization to broader task distributions. Overall, these empirical findings highlight flow-based continuous prediction and compact context modeling as effective design choices for in-context imitation learning in robotics.

\section*{Acknowledgements}
We would like to thank the anonymous reviewers for their constructive and thoughtful comments. We would like to thank Li Mi for insightful discussions. The research reported in this publication was supported by funding from King Abdullah University of Science and Technology (KAUST) - Center of Excellence for Generative AI, under award number 5940 and a gift from Google.

%
%
\bibliographystyle{splncs04}
\bibliography{main}

\clearpage
\appendix
\renewcommand{\theHsection}{supp.\arabic{section}}
\renewcommand{\theHtable}{supp.\arabic{table}}
\renewcommand{\theHfigure}{supp.\arabic{figure}}
\section*{Supplementary Material}
\setcounter{section}{0}
\renewcommand{\thesection}{S\arabic{section}}
\setcounter{table}{0}
\renewcommand{\thetable}{S\arabic{table}}
\setcounter{figure}{0}
\renewcommand{\thefigure}{S\arabic{figure}}
\newcommand\blfootnote[1]{%
  \begingroup
  \renewcommand\thefootnote{}\footnote{#1}%
  \addtocounter{footnote}{-1}%
  \endgroup
}
\newcommand{\tobj}[1]{\textcolor{blue}{\textlangle#1\textrangle}}
\newcommand{\thand}[1]{\textcolor{red}{\textlangle#1\textrangle}}

The supplementary materials are organized as follows:

\begin{itemize}[leftmargin=2em, itemsep=2pt]
    \item \textbf{Additional Implementation Details} (Section~\ref{ssec:implement}), including implementation details of ContextFlow and baselines, and detailed dataset statistics for the real-world experiments.
    \item \textbf{Supplementary Experimental Results} (Section~\ref{ssec:experiment}), including results with additional LIBERO task suites, seen task configuration performance, extended real-world experiments, and visualizations.
    \item \textbf{Supplementary Analysis} (Section~\ref{ssec:analysis}), including analysis of model behavior on the LIBERO-Object tasks suite and visualization and additional analysis of failure modes.
    \item \textbf{Discussion} (Section~\ref{ssec:discussion}).
\end{itemize}

\section{Additional Implementation Details}
\label{ssec:implement}

\subsection{Implementation Details of ContextFlow}
\label{app:model-details}

Observation images are encoded with SigLIP So400m/14~\cite{siglip} at a resolution of $224 \times 224$, producing $N_{\text{img}} = 256$ visual tokens per camera view. For each camera, we concatenate the visual tokens from the sampled demonstration frames and compress this sequence into $N_q^{\text{img}} = 32$ tokens using a shared image compressor. The state and action compressors each produce $N_q^{\text{state}} = N_q^{\text{action}} = 32$ tokens. The total number of demonstration context tokens is $N_{\text{context}} = N_{\text{cam}} \cdot N_q^{\text{img}} + N_q^{\text{state}} + N_q^{\text{action}}$: 128 for LIBERO with two cameras and 160 for ALOHA with three cameras. These compressed demonstration tokens are concatenated with the uncompressed current-observation image tokens (512 for LIBERO and 768 for ALOHA) and up to 48 text tokens, then passed to the context expert.

We set the action horizon to $H = 50$ and parameterize noise action tokens following $\pi_0$~\cite{pi0}. Both experts use multi-query attention and a Gemma-style~\cite{gemma} transformer with depth 18 and MLP dimension 4096. The context expert has embedding dimension $C_1 = 2048$ and is randomly initialized and trained in full. The action expert has embedding dimension $C_2 = 1024$ and is initialized from $\pi_0$ and fine-tuned using LoRA~\cite{hu2022lora} with rank 32 and $\alpha = 32$. The SigLIP vision encoder is also initialized from $\pi_0$ and fully fine-tuned.

\subsection{Implementation Details of Baselines}
\noindent \textbf{ICRT}~\cite{icrt}.
Following the original paper~\cite{icrt}, we fine-tune ICRT with the \textit{Llama2-Base} model, which is randomly initialized and fully fine-tuned. The sequence length for each sample is set to 512. We use an effective batch size of 8 samples, implemented as a per-device batch size of 2 with 4 gradient accumulation steps. We train the ICRT \textit{Llama2-Base} model for 5 epochs on the LIBERO seen task configurations. All remaining fine-tuning hyperparameters follow the default settings provided in the original paper~\cite{icrt}. We also experimented with training for 10 epochs, but found that 5 epochs yielded better performance. In addition, we evaluated an \textit{Llama2-7B} variant of ICRT, initialized from the pre-trained \textit{Llama2-7B} language model and fine-tuned using LoRA. Consistent with the findings in the original paper~\cite{icrt}, the \textit{Llama2-7B} variant performed worse than the \textit{Llama2-Base} model in our setting. During inference, we provide a single demonstration trajectory without subsampling as the prompt for each task.

\noindent \textbf{$\bm{\pi_0}$~\cite{pi0}.}
We follow the default settings of \textbf{$\pi_0$}~\cite{pi0} in the official implementation \footnote{\url{https://github.com/Physical-Intelligence/openpi}} and fine-tune the $\pi_0$ base model for 30K iterations with a batch size of 32 without LoRA~\cite{hu2022lora} and EMA on seen task configurations. Note that we disable \textit{extra delta transform} since the LIBERO action is of OSC-POSE and already delta.

\noindent \textbf{$\bm{\pi_0}$~\cite{pi0} FT.} For one-shot fine-tuning on each unseen task, we start from the 30k-iteration checkpoint and fine-tune on a single demonstration from that unseen task for 100 iterations, a batch size of 2 and a constant learning rate of $2.5 \times 10^{-6}$. We repeat this procedure separately for each unseen task. The reported performance for each task therefore reflects traditional one-shot learning that requires updating model weights. This contrast highlights the advantage of our in-context learning approach, which achieves comparable or superior performance without any task-specific parameter updates.

\noindent \textbf{OpenVLA-OFT}.
We follow the default training settings of LIBERO from the official implementation\footnote{\url{https://github.com/moojink/openvla-oft/blob/main/LIBERO.md}}, with the exception that we reduce the number of GPUs to four, limit the total iterations to 20,000, and use a per-GPU batch size of 8. These modifications are necessitated by our computational constraints and the exceptionally large size of the model (7B parameters). However, under this reduced configuration, we observe sub-optimal performance on seen task configurations compared to the results reported in the original OpenVLA-OFT work, which is likely attributable to our lower computational budget and modified training parameters.

\noindent \textbf{ContextAR}.
We build our in-context learning implementation on the {$\pi_0$}-FAST~\cite{pertsch2025fast} base model. Recall that $\pi_0$-FAST discretizes the normalized proprioceptive state into 256 bins and concatenates its string representation with the language instruction as part of the prefix, which is then tokenized by the PaliGemma~\cite{beyer2024paligemma} tokenizer. The action sequence is tokenized separately by the FAST tokenizer as a postfix and mapped to the final indices of the PaliGemma vocabulary. To avoid token explosion in the in-context demonstration, we implement a projection layer that maps each discrete in-context state into a single 2048-dimensional (for base model) continuous token. A parallel projection layer is applied to in-context actions. This design allows us to preserve the original training input format while effectively leveraging the pretrained model without incurring substantial architectural changes. 

\begin{table}[t]
\centering
\caption{\textbf{Seen and unseen task configurations in the real-world aloha experiments.}}
\scriptsize
\renewcommand{\arraystretch}{1.15}
\begin{minipage}{0.98\textwidth}
\begin{tabularx}{\textwidth}{@{}X@{}}
\toprule
\textbf{Pick \& Place}: ``pick up $\langle$object$\rangle$ and place it in the basket with $\langle$left/right$\rangle$ hand'' \\
\rowcolor{gray!12} \quad \textbf{Seen (14)} \quad $\langle$left$\rangle$: apple, corn, gray milk, carrot, chips \quad $\langle$right$\  rangle$: pear, orange juice, gray milk, cucumber, corn, red apple, chips, blue milk, carrot \\
\quad \textbf{Unseen (4)} \quad $\langle$left$\rangle$: pear, orange juice \quad $\langle$right$\rangle$: kiwi, banana \\
\midrule
\textbf{Pen Uncap}: ``pick up $\langle$pen$\rangle$ with $\langle$left/right$\rangle$ hand, grasp the cap with another hand and uncap it'' \\
\rowcolor{gray!12} \quad \textbf{Seen (6)} \quad $\langle$left \& right$\rangle$: gray pen, blue pen v2~ $\langle$right$\rangle$: red pen~ $\langle$left$\rangle$: blue pen \\
\quad \textbf{Unseen (1)} \quad $\langle$left$\rangle$: red pen \\
\midrule
\textbf{Put Egg in Box}: ``pick up the $\langle$object$\rangle$ with right hand, place it in the box, and close the box'' \\
\rowcolor{gray!12} \quad \textbf{Seen (1)} \quad white egg \quad\quad \textbf{Unseen (1)} \quad red egg \\
\midrule
\textbf{Extra Bimanual Task configurations} \\
\rowcolor{gray!12} \quad \textbf{Seen (4)} \quad handover, cup stack, stir, water wipe \\
\bottomrule
\end{tabularx}
\end{minipage}
\label{stab:seen_unseen_aloha}
\end{table}

\begin{table}[t]
\centering
\caption{\textbf{Training data statistics for seen task configurations.}}
\scriptsize
\renewcommand{\arraystretch}{1.05}
\begin{minipage}{0.95\textwidth}
\begin{tabular*}{\textwidth}{@{\extracolsep{\fill}}llcc@{}}
\toprule
\textbf{Task Suite} & \textbf{Task configurations} & \textbf{Episodes} & \textbf{Avg Length} \\
\midrule
\multirow{14}{*}{Pick \& Place}
 & apple / $\langle$left$\rangle$          & 25  & 200 \\
 & corn / $\langle$left$\rangle$           & 25  & 200 \\
 & gray milk / $\langle$left$\rangle$      & 25  & 200 \\
 & carrot / $\langle$left$\rangle$         & 25  & 200 \\
 & chips / $\langle$left$\rangle$          & 25  & 200 \\
 & pear / $\langle$right$\rangle$          & 50  & 250 \\
 & orange juice / $\langle$right$\rangle$  & 25  & 240 \\
 & gray milk / $\langle$right$\rangle$     & 26  & 204 \\
 & cucumber / $\langle$right$\rangle$      & 50  & 250 \\
 & corn / $\langle$right$\rangle$          & 50  & 246 \\
 & red apple / $\langle$right$\rangle$     & 25  & 200 \\
 & chips / $\langle$right$\rangle$         & 25  & 200 \\
 & blue milk / $\langle$right$\rangle$     & 76  & 267 \\
 & carrot / $\langle$right$\rangle$        & 75  & 265 \\
\cmidrule{2-4}
 & \textit{Subtotal}                       & \textit{527} & \textit{235} \\
\midrule
\multirow{6}{*}{Pen Uncap}
 & gray pen / $\langle$right$\rangle$      & 143 & 604 \\
 & gray pen / $\langle$left$\rangle$       & 51  & 500 \\
 & red pen / $\langle$right$\rangle$        & 69  & 497 \\
 & blue pen / $\langle$left$\rangle$       & 25  & 500 \\
 & blue pen v2 / $\langle$left$\rangle$    & 25  & 500 \\
 & blue pen v2 / $\langle$right$\rangle$   & 25  & 500 \\
\cmidrule{2-4}
 & \textit{Subtotal}                       & \textit{338} & \textit{543} \\
\midrule
Put Egg in Box
 & white egg                               & 199 & 735 \\
\cmidrule{2-4}
 & \textit{Subtotal}                       & \textit{199} & \textit{735} \\
\midrule
\multirow{4}{*}{\shortstack[l]{Extra Bimanual\\(train only)}}
 & handover                                & 104 & 642 \\
 & cup stack                               & 50  & 400 \\
 & stir                                    & 50  & 350 \\
 & water wipe                              & 50  & 500 \\
\cmidrule{2-4}
 & \textit{Subtotal}                       & \textit{254} & \textit{509} \\
\midrule
\multicolumn{2}{@{}l}{\textbf{Total (3 main suites)}} & \textbf{1,064} & -- \\
\multicolumn{2}{@{}l}{\textbf{Total (all)}}           & \textbf{1,318} & -- \\
\bottomrule
\end{tabular*}
\end{minipage}
\label{stab:training_data_stats}
\end{table}

\subsection{Details of Real-World Experiments}
\label{sec:real-world-tasks}
We conducted real-world experiments on a mobile ALOHA robot platform. The scene consists of a table and toy objects (e.g., toy fruits, vegetables, eggs) and real pens. The setup includes one high camera and two wrist-mounted cameras, and demonstrations are collected via human teleoperation. We collected a dataset that includes both single arm task configurations and fine-grained bimanual manipulation task configurations. Detailed task configurations are shown in Tab.~\ref{stab:seen_unseen_aloha}. The statistics of training data are shown in Tab.~\ref{stab:training_data_stats}.
Pick Up and Place tasks follow the instruction template \texttt{``Pick up the \textless object\textgreater{} and place it in the basket with the \textless left/right\textgreater{} hand''.} There are two possible basket locations: (i) in front of the midpoint between the two arms, in which case the task is executed with the right hand, and (ii) in front of the left arm, in which case the task is executed with the left hand. Across task configurations, we vary the positions of both the objects and the basket to increase task diversity. Pen Uncap tasks follow the instruction template \texttt{``Pick up the \textless pen\textgreater{} with the \textless left/right\textgreater{} hand, grasp the cap with the other hand and uncap it''.} There are 4 different kinds of pens (gray, red, blue, and blue v2), varying in shape, color, and force properties required for uncapping. For each pen, there are two motion trajectory variants depending on which hand is used to pick up the pen first; the other hand then grasps and removes the cap, making this a bimanual coordination task. Put Egg in Box tasks follow the instruction template \texttt{``Pick up the \textless object\textgreater{} with right hand, place it in the box, and close box''.} This is another fine-grained bimanual task that requires the robot to pick up the egg, carefully place it into the box, and then close the lid using both hands. Training data includes putting a white egg in the box and closing it, while the unseen test task uses a red egg to evaluate generalization to novel object appearance. In addition, we include 4 extra bimanual tasks (handover, cup stack, stir, and water wipe) with 254 demonstrations for training only. These tasks further enrich the training distribution with diverse bimanual manipulation behaviors.

\section{Supplementary Experimental Results}
\label{ssec:experiment}

\begin{table}[h]
\centering
\caption{\textbf{Success rates on seen and unseen task configurations across all task suites in LIBERO.}}
\begin{minipage}{0.98\textwidth}
\scriptsize
\begin{tabular*}{\textwidth}{@{\extracolsep{\fill}}l ccccc ccccc@{}}
\toprule
\multirow{2}{*}{\textbf{Model}} &
\multicolumn{5}{c}{\textbf{Seen Task Configurations}} &
\multicolumn{5}{c}{\textbf{Unseen Task Configurations}} \\
\cmidrule{2-6} \cmidrule{7-11}
& Spatial & Object & Goal & 10 & Avg. &
Spatial & Object & Goal & 10 & Avg. \\
\midrule
$\pi_0$~\cite{pi0} &
\textbf{96.0} & 97.0 & \textbf{94.0} & \textbf{87.0} & \textbf{93.5} &
26.0 & 63.0 & 0.0 & 0.0 & 22.3 \\
ICRT~\cite{icrt} &
85.0 & \textbf{99.0} & 90.0 & 65.0 & 84.8 &
27.0 & 50.0 & \textbf{1.0} & 0.0 & 19.5 \\
ContextFlow (Ours) &
91.0 & 97.0 & 90.0 & 82.0 & 90.0 &
\textbf{64.0} & \textbf{83.0} & 0.0 & 0.0 & \textbf{36.8} \\
\bottomrule
\end{tabular*}
\end{minipage}
\label{stab:seen_unseen_comparison_final}
\end{table}

\subsection{Results on unseen and seen tasks across all task suites}

We compare our method against $\pi_0$~\cite{pi0} and ICRT~\cite{icrt} on all task configurations, including both seen and unseen task configurations across four task suites, as shown in Tab.~\ref{stab:seen_unseen_comparison_final}. Although ContextFlow is designed for \textit{unseen} task configurations, it achieves performance comparable to $\pi_0$ on the seen task configurations. We observe that all models struggle on the Goal and Long-Horizon-10 \textit{unseen} task configurations, likely due to the larger task variations in these suites. These results highlight the limitations of current in-context imitation learning methods. 

Table~\ref{stab:unseen_tasks} lists the two held-out testing tasks in each of the four LIBERO suites. These tasks are excluded from training.
Details of all task indexes and corresponding language descriptions are available on Hugging Face\footnote{\url{https://huggingface.co/datasets/physical-intelligence/libero/blob/main/meta/tasks.jsonl}}.

\begin{table}[h]
\centering
\renewcommand{\arraystretch}{1.3}
\caption{\textbf{Unseen task configurations in LIBERO.}}
\resizebox{\linewidth}{!}{
\begin{tabular}{ccc}
\toprule
\textbf{Task Suites} & \textbf{Task Indexes} & \textbf{Language Descriptions} \\
\midrule
\multirow{2}{*}{Spatial} & 35 & Pick up the black bowl on the cookie box and place it on the plate \\
                       & 36 & Pick up the black bowl next to the plate and place it on the plate \\
\cline{1-3}
\multirow{2}{*}{Object} & 25 & Pick up the milk and place it in the basket \\
                      & 28 & Pick up the tomato sauce and place it in the basket \\
\cline{1-3}
\multirow{2}{*}{Goal} & 10 & Put the bowl on the plate \\
                    & 17 & Put the bowl on the stove \\
\cline{1-3}
\multirow{2}{*}{Long-horizon 10} & 1 & Put the white mug on the plate and put the chocolate pudding to the right of the plate \\
                               & 5 & Put both the alphabet soup and the tomato sauce in the basket \\
\bottomrule
\end{tabular}
}
\label{stab:unseen_tasks}
\end{table}

\subsection{Comparison on Real-world experiments}
We additionally evaluate $\pi_0$~\cite{pi0} in the real-world experiments. As shown in Tab.~\ref{stab:real_world}, ContextFlow demonstrates stronger generalization to unseen task configurations compared to both baselines. We observe that success rates are influenced by both the intrinsic task difficulty and the gap between seen and unseen task configurations. For single-armed pick-and-place tasks, which are short-horizon and relatively simple, the unseen test objects and positions differ substantially from those seen during training. In this setting, $\pi_0$ achieves moderate success rates but remains significantly below ContextFlow. For the put-egg-in-box task, although it is a long-horizon fine-grained bimanual task, the difference between seen and unseen configurations is minimal (white egg vs.\ red egg), and $\pi_0$ performs well. In contrast, the uncap-red-pen task combines fine-grained bimanual coordination with a larger task configuration gap between seen and unseen. Here, $\pi_0$ fails entirely, while ContextFlow still maintains a reasonable success rate.

\subsection{Visualization of ContextFlow Rollout}
\begin{figure}[t]
\centering
\includegraphics[width=0.98\linewidth]{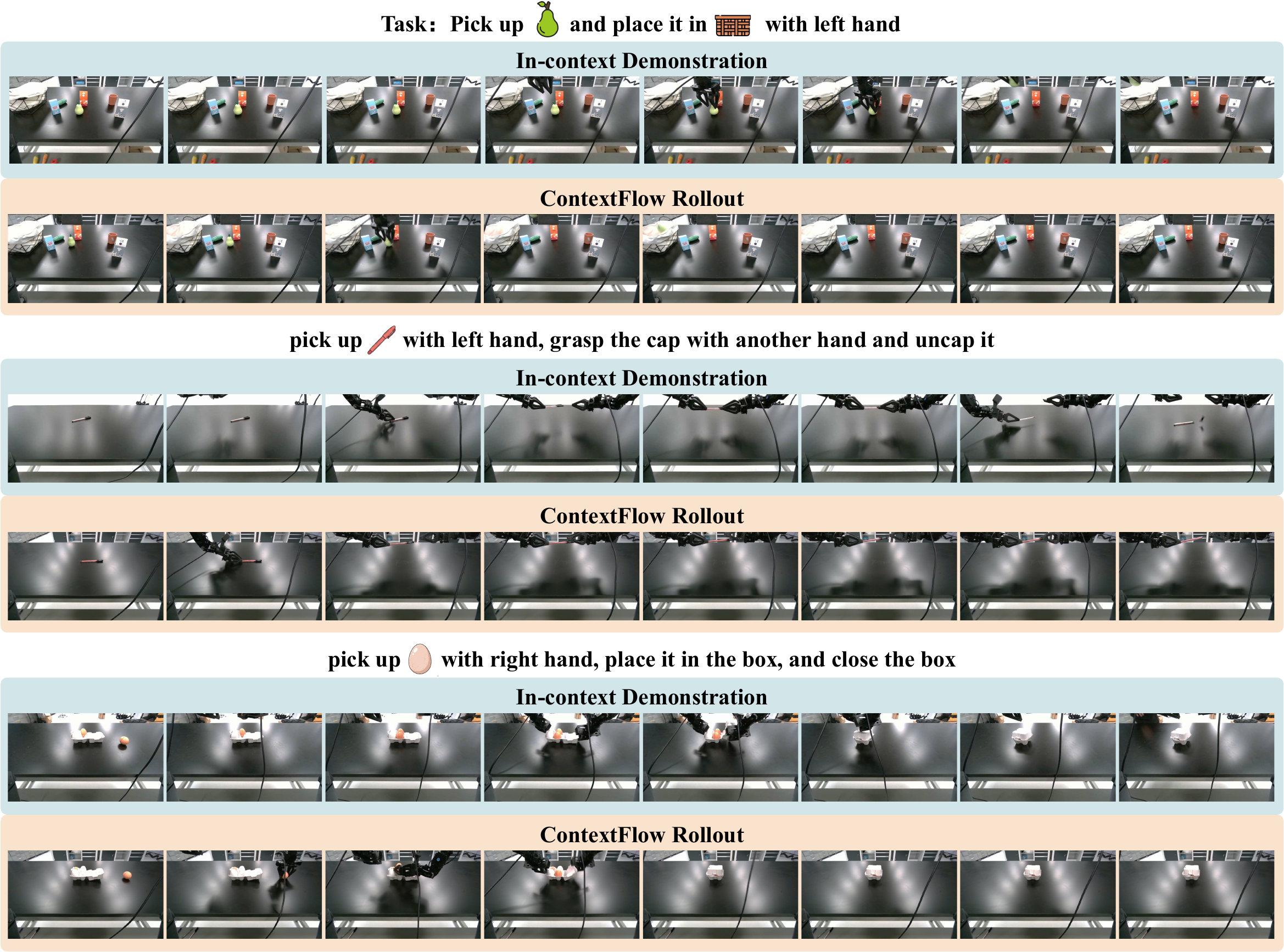}
    \caption{\textbf{Rollout of ContextFlow in real-world experiments.} We show the in-context demonstrations and the corresponding ContextFlow rollouts from high camera.}
    \label{sfig:contextflow_rollout_real}
\end{figure}

\begin{figure}[t]
\centering
\includegraphics[width=0.85\linewidth]{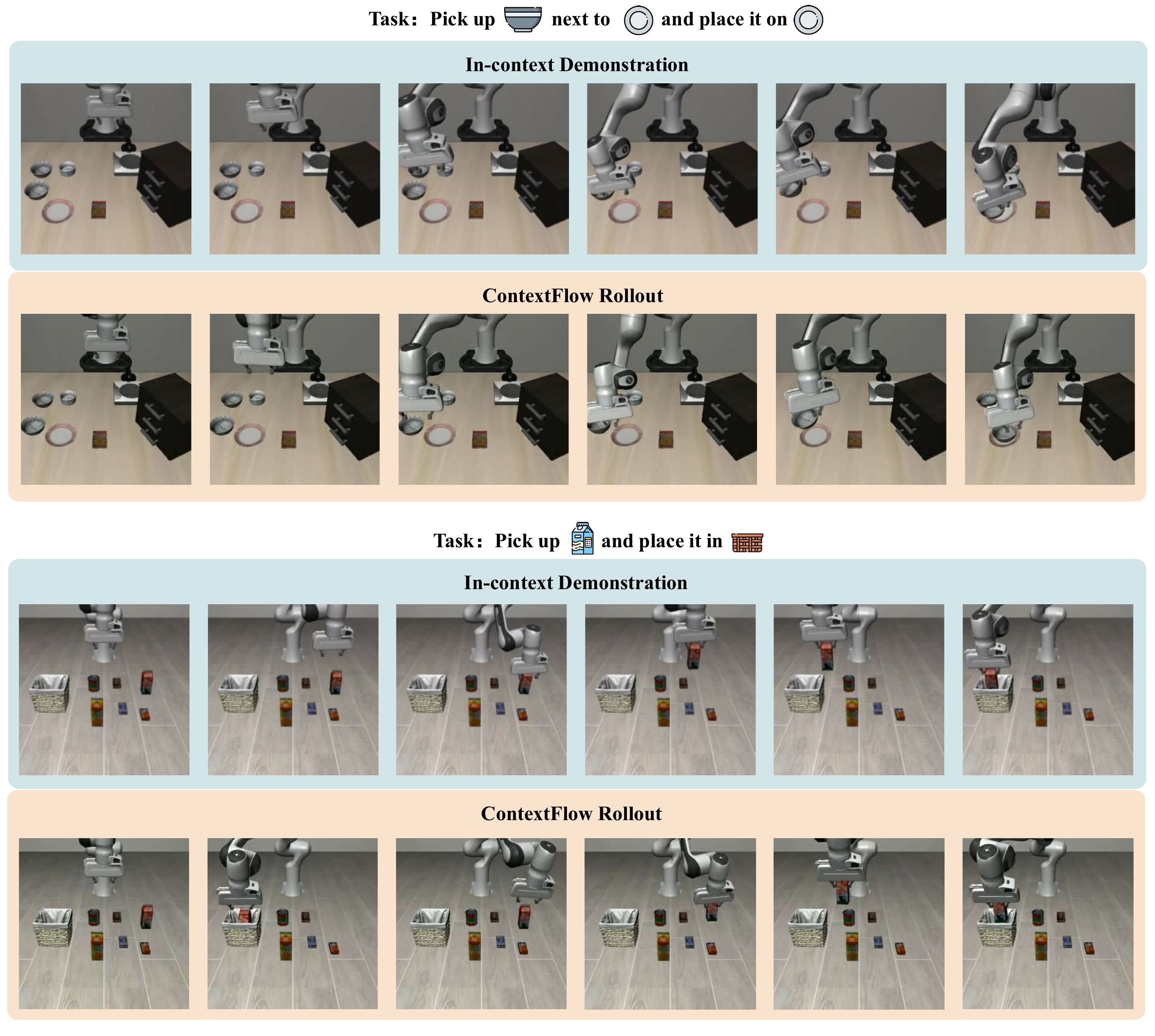}
    \caption{\textbf{Rollout of ContextFlow in LIBERO simulation.} We show the in-context demonstrations and the corresponding ContextFlow rollouts from both agent and wrist views. ``Pick up black bowl next to the plate and place it on the plate'' belongs to LIBERO Spatial task suite and ``Pick up the milk and place it in the basket'' belongs to LIBERO Object task suite.}
    \label{sfig:contextflow_rollout_sim}
\end{figure}

\begin{figure}[t]
\centering
\includegraphics[width=0.98\linewidth]{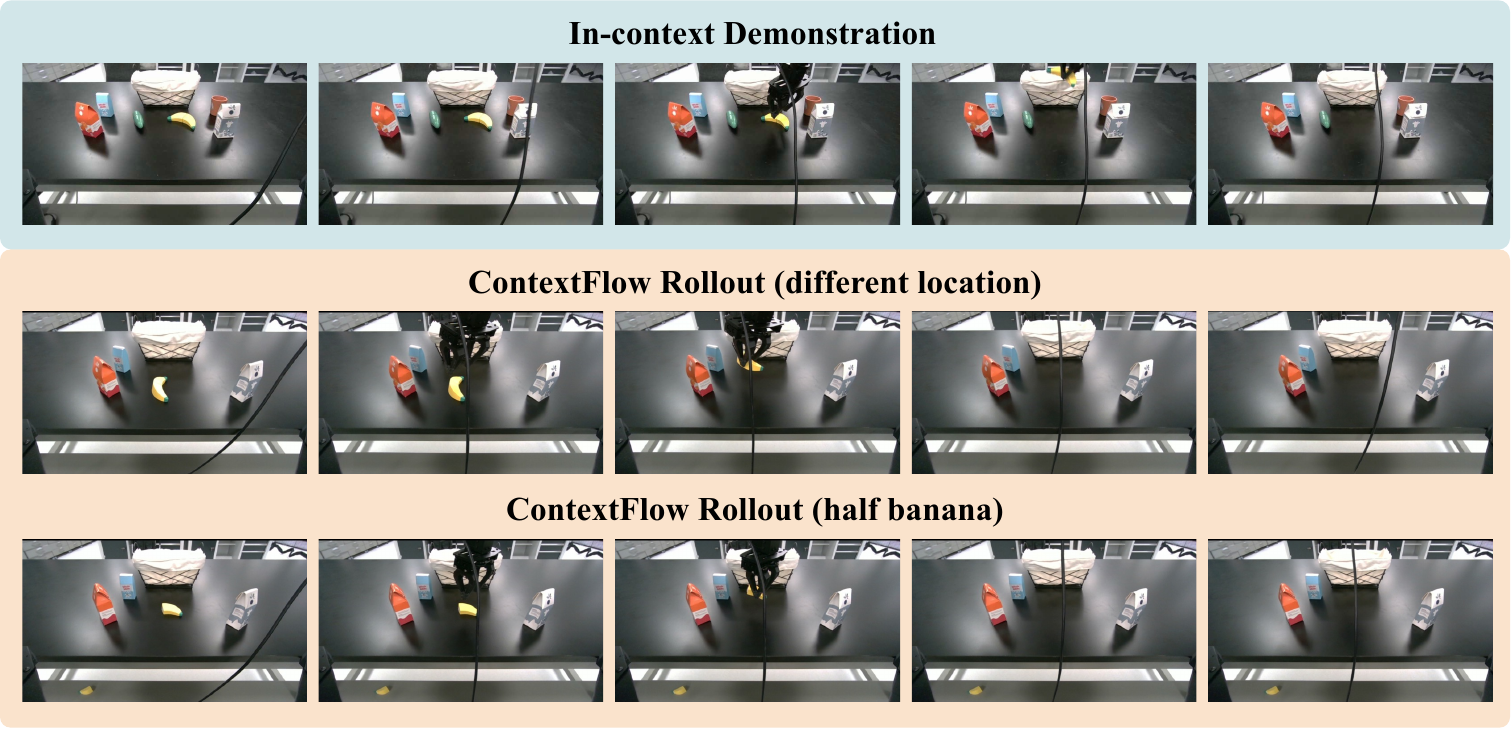}
    \caption{\textbf{Generalization of ContextFlow under distribution shifts from the in-context demonstration.} The first row shows the in-context demonstration. The second row shows the rollout where the banana is placed at a different location and orientation. The third row shows the rollout where the banana is cut in half, introducing a novel object appearance.}
    \label{sfig:banana_diff}
\end{figure}

In Fig.~\ref{sfig:contextflow_rollout_real} and Fig.~\ref{sfig:contextflow_rollout_sim}, we visualize an in-context demonstration and the corresponding ContextFlow rollout in the real world and in simulation, respectively. Note that the initial states of the in-context demonstration and the test scene are similar but not identical. For example, in the pick-and-place pear task, the pear location differs slightly, and in the pen uncap task, the pen location and orientation vary between the demonstration and test scene. Additional rollout examples are provided in the supplementary videos.

\noindent\textbf{Robustness to discrepancies between in-context demonstrations and test scenes.} In practice, the initial state of the test scene is rarely identical to that of the in-context demonstration. We further evaluate ContextFlow's robustness by intentionally increasing such discrepancies, with qualitative examples shown in Fig.~\ref{sfig:banana_diff}. The first example involves a spatial difference, where the banana is placed at a significantly different location from the demonstration. The second example involves an object-level difference, where the banana is cut in half and only one piece is used for testing. In both cases, ContextFlow successfully completes the task despite the discrepancies, demonstrating its ability to generalize across variations in initial conditions.

\subsection{Computation cost for training and inference}
We reported A100 GPU hours and inference throughput (Hz.) in Tab.~\ref{tab:speed}. We followed OpenVLA-OFT~\cite{kim2025fine} for the computation of throughput. We can see that flow matching models are usually faster than the autoregressive models due to the prediction of action chunk in parallel. ContextFlow-Plain is fast because it uses only 2 images. However, increasing the number of image tokens in ContextFlow-Plain leads to degraded performance. This suggests that without an image compressor (used in ContextFlow), directly attending to a large number of tokens is not only computationally expensive but also makes it difficult to attend to the relevant tokens for in-context imitation learning.

\begin{table}[t]
\centering
\caption{\textbf{Success rates on unseen task configurations on a real-world robot.} See Sec.~\ref{sec:real-world-tasks} for task instruction templates. Elements in different task configurations are listed in the table. Note that the pen uncap task here uses the left hand to pick up the pen (i.e., ``pick up the red pen with left hand, grasp the cap with the other hand and uncap it''.)}
\label{stab:real_world}
\begin{minipage}{0.99\textwidth}
{\scriptsize
\setlength{\tabcolsep}{2.5pt}
\begin{tabular*}{\textwidth}{@{\extracolsep{\fill}}l|cccc|cc|c}
\toprule
\multirow{3}{*}{\textbf{Model}} &
\multicolumn{4}{c|}{\textbf{Single-armed Pick-and-place tasks}} &
\multicolumn{2}{c|}{\textbf{Bimanual tasks}} &
\multirow{3}{*}{\textbf{Avg.}} \\
\cmidrule{2-5} \cmidrule{6-7}
& \multicolumn{2}{c}{\textbf{Left hand}} &
\multicolumn{2}{c|}{\textbf{Right hand}} &
\textbf{Uncap} & \textbf{Put in box} & \\
& \includegraphics[height=0.9em]{figures/Fig_Icon_Pear.png} Pear
& \includegraphics[height=0.9em]{figures/Fig_Icon_Orange.png} Orange
& \includegraphics[height=0.9em]{figures/Fig_Icon_Bananas.png} Banana
& \includegraphics[height=0.9em]{figures/Fig_Icon_Kiwi.png} Kiwi
& \includegraphics[height=0.9em]{figures/Fig_Icon_Red_Pen.png} Red Pen
& \includegraphics[height=0.9em]{figures/Fig_Icon_Red_Egg.png} Red Egg & \\
\midrule
ICRT~\cite{icrt} & \textbf{2/10} & 0/10 & 0/10 & 0/10 & 0/10 & 0/10 & 3.3 \\
$\pi_0$~\cite{pi0}              & \textbf{2/10} & 2/10 & 3/10 & 2/10 & 0/10 & \textbf{7/10} & 26.7 \\
ContextFlow (Ours)     & \textbf{2/10} & \textbf{5/10} & \textbf{4/10} & \textbf{4/10} & \textbf{4/10} & 6/10 & \textbf{41.7} \\
\bottomrule
\end{tabular*}
}
\end{minipage}
\end{table}

\begin{table}[!ht]
\centering
\caption{Training and inference computational costs of different models. We report the training time in A100 GPU hours and the inference throughput (Hz) measured on a single A100 GPU.}
\begin{minipage}{0.65\textwidth}
\scriptsize
\begin{tabular*}{\textwidth}{@{\extracolsep{\fill}}l|cc}
\toprule
\textbf{Model} & \textbf{Training (Hours)} $\downarrow$ & \textbf{Inference (Hz)} $\uparrow$ \\
\midrule
OpenVLA-OFT~\cite{kim2025fine} & 30 & 77.8 \\
$\pi_0$~\cite{pi0} & 21 & 569.0 \\
ICRT~\cite{icrt} & 16 & 44.3 \\
ContextAR & 65 & 3.7 \\
ContextFlow-Plain & 21 & 310.2 \\
ContextFlow & 40 & 130.6 \\
\bottomrule
\end{tabular*}
\end{minipage}
\label{tab:speed}
\end{table}

\section{Supplementary Analysis}
\label{ssec:analysis}

\subsection{Model Behavior Analysis on LIBERO-Object}

\begin{figure}[t]
\centering
\includegraphics[width=0.98\linewidth]{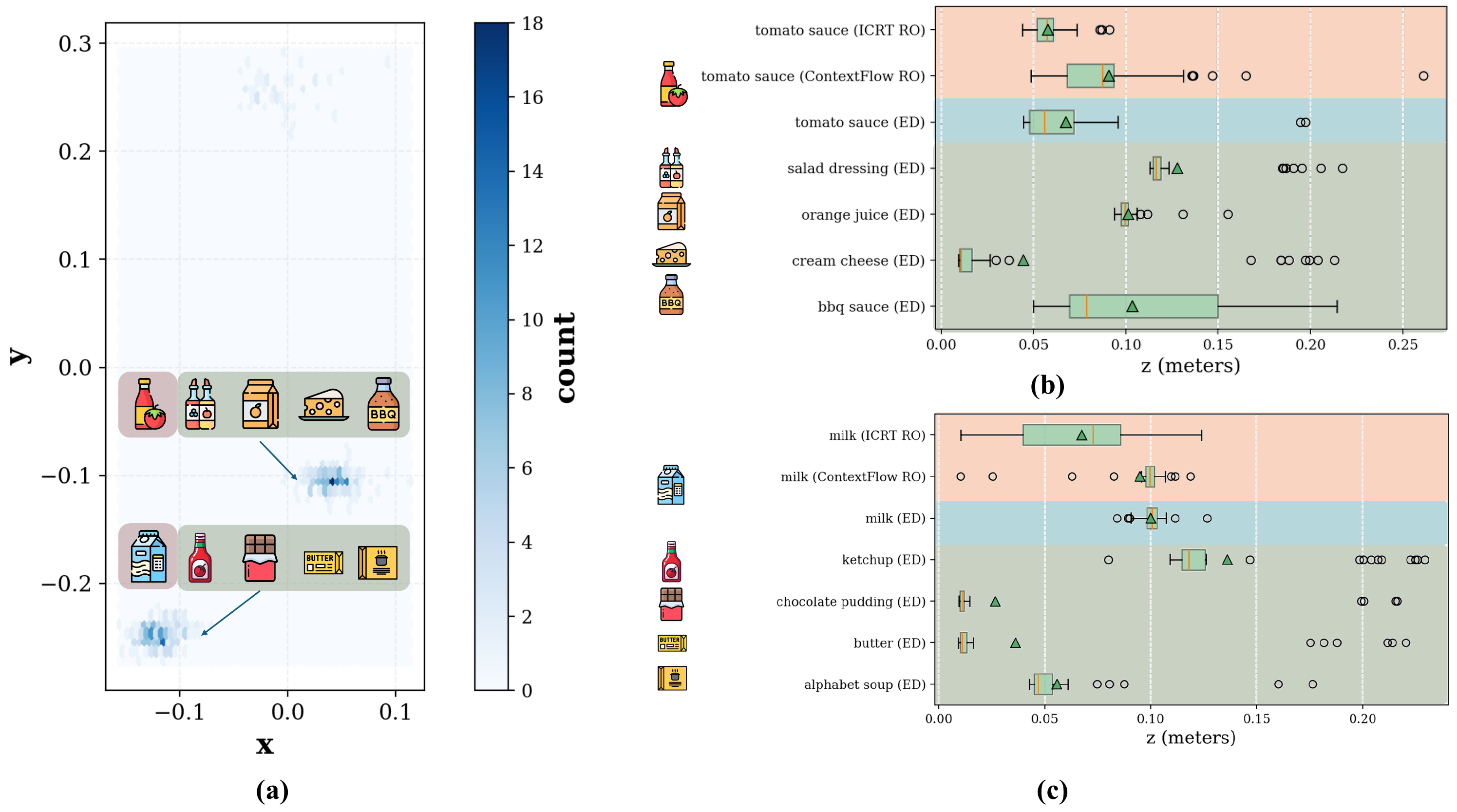}
    \caption{\textbf{Grasp position analysis of the LIBERO-object task suite.} (a) XY distribution of grasp positions across all demonstrations, revealing two distinct spatial clusters. (b) Z-coordinate distributions for task configurations belonging to the cluster that contains the unseen \textit{tomato sauce} task, whose object height is close to the training average. (c) Z-coordinate distributions for task configurations belonging to the cluster that contains the unseen \textit{milk} task, whose object height is notably higher than the training average. In (b) and (c), green triangles mark the mean Z-coordinate across multiple demonstrations, yellow vertical lines denote the median, and dots indicate outliers. Tasks configurations in the lower green region correspond to \textit{seen} expert demonstrations (ED); the middle blue region corresponds to \textit{unseen} task configurations in LIBERO-Object; and the upper red region corresponds to ICRT and ContextFlow inference rollouts (RO). We observe that when the height of an \textit{unseen} object differs from the training average, ICRT rollouts tend to follow the average grasp position and fail to match the unseen object’s height, whereas our proposed ContextFlow generalizes well and aligns its grasp height with the unseen objects.}
    \label{sfig:object_z}
\end{figure}

We approximate each 3D grasp position as the end-effector position at the time step with the largest frame-to-frame change in gripper openness. The XY distribution of these grasp positions across all LIBERO Object episodes exhibits two clear spatial clusters (Fig.~\ref{sfig:object_z}~(a)), so we further analyze the Z-dimension within each cluster. Specifically, we examine the Z-coordinate distributions of grasp positions for seen and unseen expert demonstrations (with green and blue background, respectively) and for inference rollouts (RO) of ICRT and ContextFlow (with orange background). In the cluster shown in Fig.~\ref{sfig:object_z}~(b), the average grasp height in the \textit{seen} task configurations is close to that of the \textit{unseen tomato sauce} object, and the grasp-height distributions of both ICRT RO and ContextFlow RO align well with the expert demonstrations, leading to comparatively high success rates. In contrast, in the cluster shown in Fig.~\ref{sfig:object_z}~(c), most training objects (such as ketchup and chocolate) have relatively low grasp points, lowering the average grasp height, whereas the unseen \textit{milk} object is positioned noticeably higher. This mismatch between the seen-task average and the \textit{milk} object makes the task challenging: ICRT achieves a 4.0\% success rate, while ContextFlow reaches 76\%. Consistent with this difference, the ICRT grasp-height distribution remains concentrated around the seen-task average, which is associated with a higher rate of grasp failures, whereas ContextFlow shifts its grasp heights upward and better aligns with the unseen object. Overall, this analysis indicates that grasp-height variation within a spatial cluster can substantially influence performance, and that ContextFlow adjusts its grasp height more flexibly to accommodate unseen object geometry under these conditions.

\begin{figure}[t]
\centering
\includegraphics[width=0.64\linewidth]{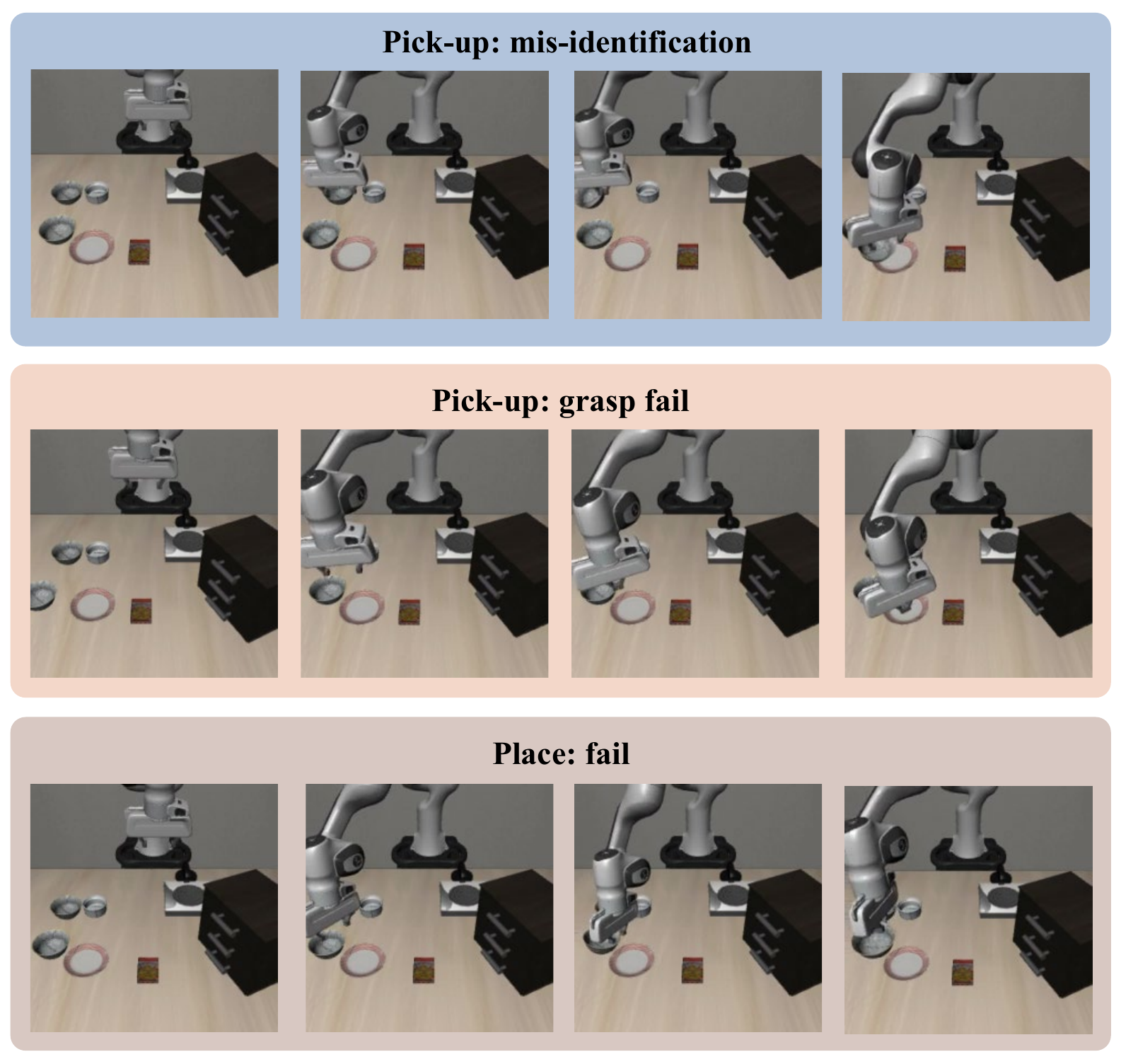}
    \caption{\textbf{Visualization of three types of failure modes.} The task for the visualization is ``pick up the black bowl next to the plate and place it on the plate.''
}
    \label{sfig:failure_s}
\end{figure}

\subsection{Additional Failure Mode Analysis}

\noindent \textbf{Visualization of Failure Modes.} In Fig 5. \textit{Failure mode analysis} in the main paper, we provide failure mode analysis of $S_2$ ``Pick up black bowl next to the plate and place it on the plate task''. We further visualize three types of failure modes in Fig.~\ref{sfig:failure_s}. The first is \textit{mis-identification}, where the model acts on the wrong object despite being given the correct prompt. This often arises when multiple visually or semantically similar objects are in close proximity, creating ambiguity in target selection. For example, the gripper tries to grasp the plate instead of the bowl. The second failure mode is \textit{grasp failure}, where the model identifies the correct object but the gripper fails to establish a successful grasp. This type of failure may be related to inaccurate action prediction. The third is \textit{place failure}, where the robot grasps the object correctly but fails to place it at the target location.

\subsection{Attention in Image Compressor}
\begin{figure}[t]
\centering
\includegraphics[width=0.92\linewidth]{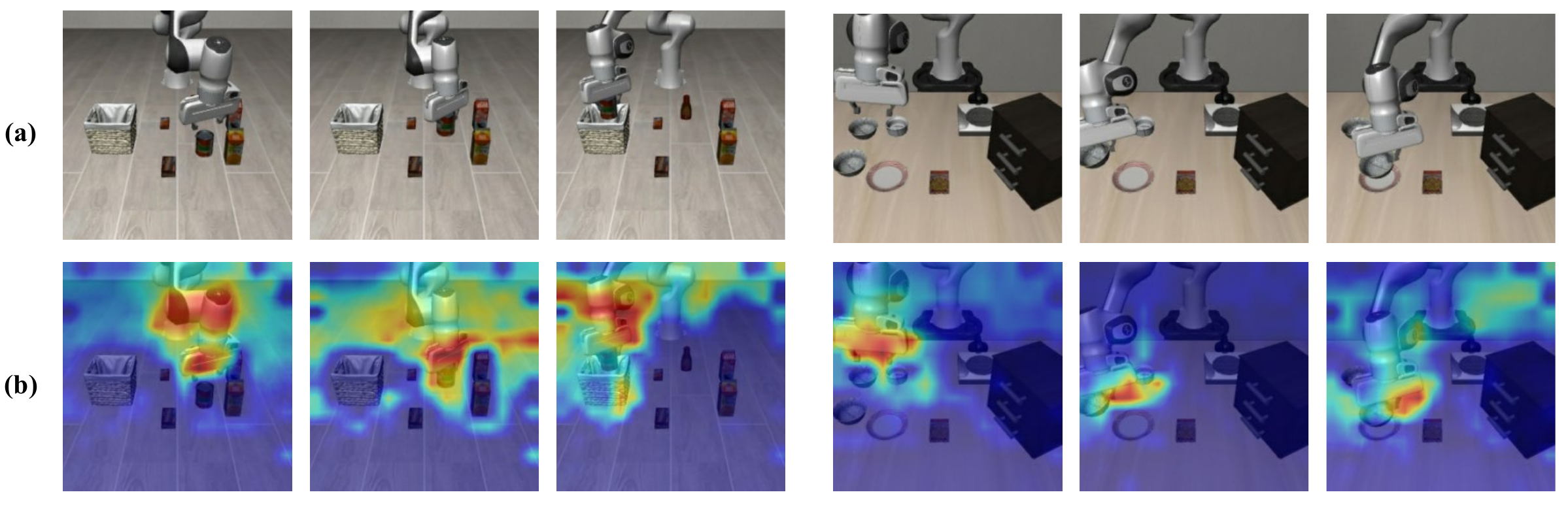}
    \caption{\textbf{Visualization of attention in image compressor on unseen task configurations.} (a) Frames in In-context Demonstration (b) Image compressor attention. The left shows the task ``pick up the tomato sauce and place it in the basket'', while the right shows the task ``pick up the black bowl next to the plate and place it on the plate''.
}
    \label{sfig:attention}
\end{figure}

To understand what the image compressor learns, we have visualized the attention of learnable queries for task ``pick up the black bowl on the cookie box and place it on the plate'' in main paper. Here, we provide visualization for other unseen task configurations in Fig.~\ref{sfig:attention}. Similarly, we notice attention changes along task phases.

\section{Discussion}
\label{ssec:discussion}
Our experiments show that flow-matching policies achieve higher accuracy and more robust generalization than autoregressive models in the in-context imitation setting, particularly on unseen task configurations where distribution shift exacerbates compounding prediction errors. By predicting continuous action chunks in parallel, flow matching avoids the strict token-by-token dependency of autoregressive models and thus reduces error accumulation over long horizons. On the other hand, recent work such as $\pi_0$-Fast~\cite{pertsch2025fast} suggests that autoregressive heads can be trained more efficiently than flow-based ones, benefiting from a simpler next-token prediction objective and faster convergence. This trade-off points to a promising hybrid direction: first train an in-context imitation model with an autoregressive objective for efficiency, then fine-tune a flow-matching head for improved robustness and accuracy, such as $\pi_{0.5}$~\cite{pi05}. Exploring such staged or joint AR–flow training schemes is an interesting avenue for future work.

\newpage

\clearpage

\end{document}